\documentclass{article}

\usepackage{microtype}
\usepackage{graphicx}
\usepackage{subcaption}
\usepackage{booktabs} 
\usepackage{amsmath}
\usepackage{enumitem}
\usepackage{mathtools}
\usepackage{amssymb}
\usepackage{amsthm}
\usepackage{stmaryrd}

\usepackage{hyperref}

\DeclareMathOperator*{\argmax}{\mathrm{argmax}}
\DeclareMathOperator*{\argmin}{\mathrm{argmin}}

\DeclareMathOperator*{\KL}{\mathrm{KL}}
\newcommand*\diff{\mathop{}\!\mathrm{d}}

\usepackage[accepted]{icml2020}

\icmltitlerunning{A Distributional View on Multi-Objective Policy Optimization}
 
\newcommand{\eref}[1]{(\ref{#1})}
\newcommand{\sref}[1]{Sec. \ref{#1}}
\newcommand{\figref}[1]{Fig. \ref{#1}}

\newcommand{\tableref}[1]{Table \ref{#1}}

\newcommand\Tstrut{\rule{0pt}{2.6ex}}         

\newcommand{\plh}{{\mkern-0.1mu\times\mkern-0.1mu}}

\newcommand{\prg}[1]{\noindent\textbf{#1. }}
\renewcommand{\vec}[1]{\mathbf{#1}}

\newcommand{\bs}[1]{\boldsymbol{#1}}

\renewcommand{\a}[0]			{ {\bs{\theta}} }

\renewcommand{\vec}[1]			{ \bs{#1} }

\DeclareUrlCommand\ULurl{%
  \renewcommand\UrlLeft{\uline\bgroup}%
  \renewcommand\UrlRight{\egroup}}
  
\DeclareMathOperator\atanh{atanh}

\begin{document}

\twocolumn[
\icmltitle{A Distributional View on Multi-Objective Policy Optimization} 
\icmlsetsymbol{equal}{*}

\begin{icmlauthorlist}
\icmlauthor{Abbas Abdolmaleki}{equal,deep}
\icmlauthor{Sandy H. Huang}{equal,deep}
\icmlauthor{Leonard Hasenclever}{deep}
\icmlauthor{Michael Neunert}{deep}
\icmlauthor{H. Francis Song}{deep}
\icmlauthor{Martina Zambelli}{deep}
\icmlauthor{Murilo F. Martins}{deep}
\icmlauthor{Nicolas Heess}{deep}
\icmlauthor{Raia Hadsell}{deep}
\icmlauthor{Martin Riedmiller}{deep}
\end{icmlauthorlist}

\icmlaffiliation{deep}{DeepMind}

\icmlcorrespondingauthor{Abbas Abdolmaleki}{aabdolmaleki@google.com}
\icmlcorrespondingauthor{Sandy H. Huang}{shhuang@google.com}

\icmlkeywords{multi-objective reinforcement learning, MORL}

\vskip 0.3in
]

\printAffiliationsAndNotice{\icmlEqualContribution} 
\begin{abstract}
Many real-world problems require trading off multiple competing objectives. However, these objectives are often in different units and/or scales, which can make it challenging for practitioners to express numerical preferences over objectives in their native units. In this paper we propose a novel algorithm for multi-objective reinforcement learning that enables setting desired preferences for objectives in a scale-invariant way. We propose to learn an action distribution for each objective, and we use supervised learning to fit a parametric policy to a combination of these distributions. We demonstrate the effectiveness of our approach on challenging high-dimensional real and simulated robotics tasks, and show that setting different preferences in our framework allows us to trace out the space of nondominated solutions.
\end{abstract}

\section{Introduction}

Reinforcement learning (RL) algorithms do an excellent job at training policies to optimize a single scalar reward function. Recent advances in deep RL have made it possible to train policies that exceed human-level performance on Atari \cite{mnih2015human} and Go \cite{Silver_2016}, perform complex robotic manipulation tasks \cite{Zeng_2019}, learn agile locomotion \cite{Tan_2018}, and even obtain reward in unanticipated ways \cite{Amodei_2016}.

However, many real-world tasks involve \emph{multiple}, possibly competing, objectives. For instance, choosing a financial portfolio requires trading off between risk and return; controlling energy systems requires trading off performance and cost; and autonomous cars must trade off fuel costs, efficiency, and safety. Multi-objective reinforcement learning (MORL) algorithms aim to tackle such problems \cite{Roijers_2013,Liu_2015}. A common approach is \emph{scalarization}: based on \emph{preferences} across objectives, transform the multi-objective reward vector into a single scalar reward (e.g., by taking a convex combination), and then use standard RL to optimize this scalar reward.

\begin{figure}[t!]
    \centering
    \begin{subfigure}{0.12\textwidth}
      \centering
      \includegraphics[width=0.96\linewidth]{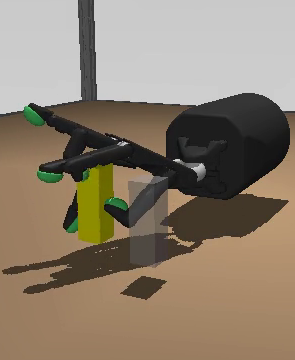}  
    \end{subfigure}%
    \begin{subfigure}{0.12\textwidth}
      \centering
      \includegraphics[width=0.96\linewidth]{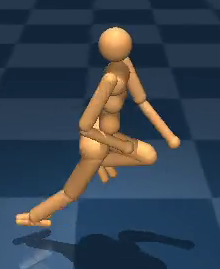}
    \end{subfigure}%
    \begin{subfigure}{0.12\textwidth}
      \centering
      \includegraphics[width=0.96\linewidth]{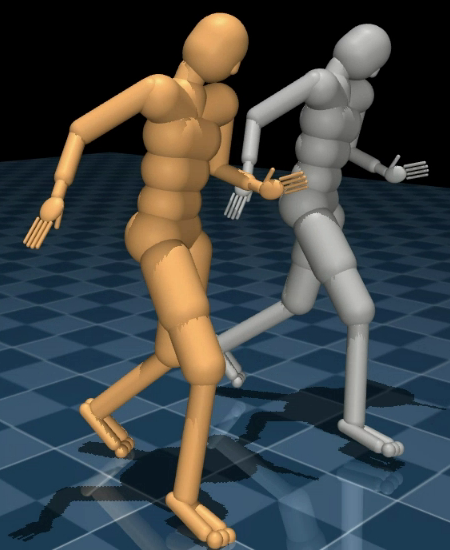}
    \end{subfigure}%
    \begin{subfigure}{0.12\textwidth}
      \centering
      \includegraphics[width=0.96\linewidth]{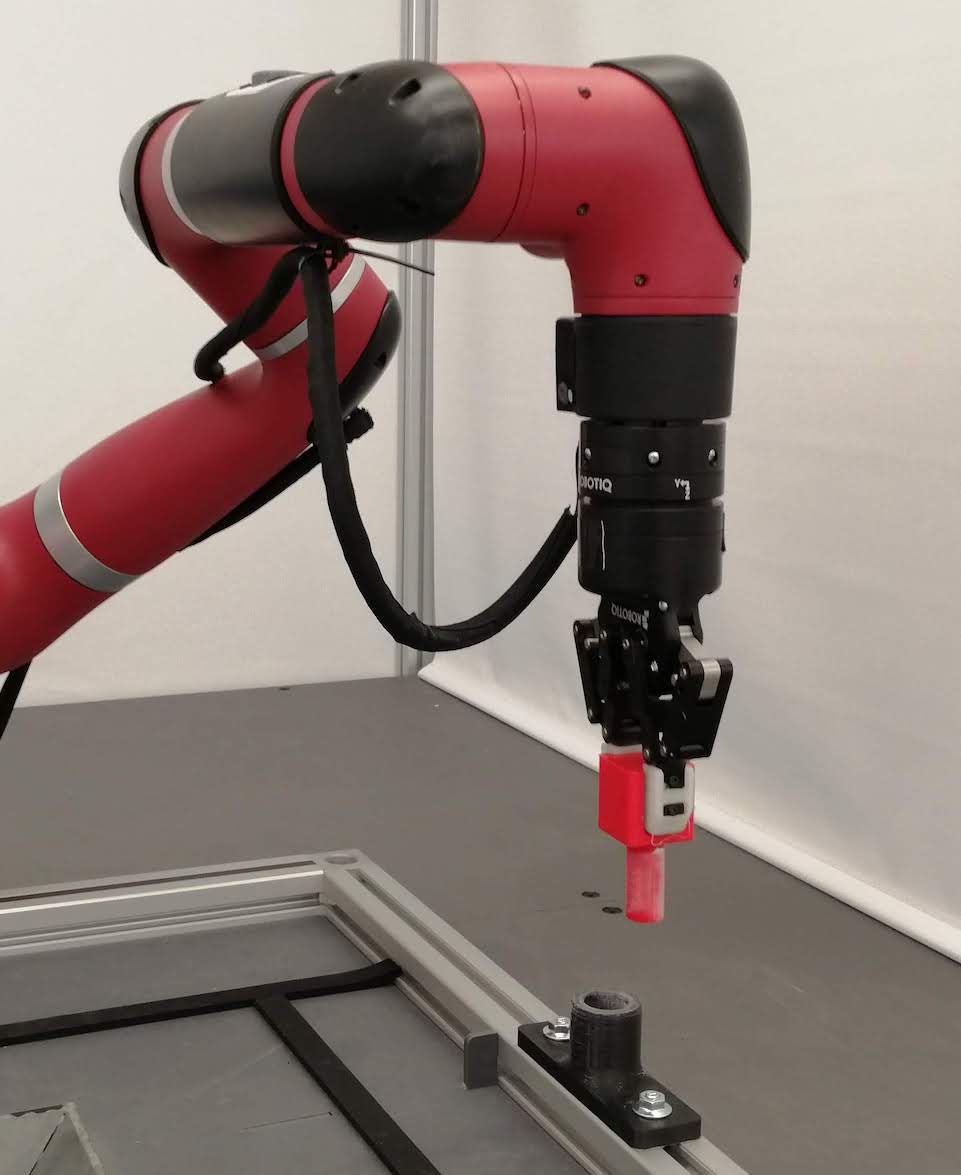}
    \end{subfigure}
    \caption{We demonstrate our approach in four complex continuous control domains, in simulation and in the real world. Videos are at \url{http://sites.google.com/view/mo-mpo}.}
    \label{fig:domains}
\end{figure}

It is tricky, though, for practitioners to pick the appropriate scalarization for a desired preference across objectives, because often objectives are defined in different units and/or scales. For instance, suppose we want an agent to complete a task while minimizing energy usage and mechanical wear-and-tear. Task completion may correspond to a sparse reward or to the number of square feet a vacuuming robot has cleaned, and reducing energy usage and mechanical wear-and-tear could be enforced by penalties on power consumption (in kWh) and actuator efforts (in N or Nm), respectively. Practitioners would need to resort to using trial and error to select a scalarization that ensures the agent prioritizes actually doing the task (and thus being useful) over saving energy.

Motivated by this, we propose a scale-invariant approach for encoding preferences, derived from the RL-as-inference perspective. Instead of choosing a scalarization, practitioners set a constraint per objective. Based on these constraints, we learn an action distribution per objective that improves on the current policy. Then, to obtain a single updated policy that makes these trade-offs, we use supervised learning to fit a policy to the combination of these action distributions. The constraints control the influence of each objective on the policy, by constraining the KL-divergence between each objective-specific distribution and the current policy. The higher the constraint value, the more influence the objective has. Thus, a desired preference over objectives can be encoded as the relative magnitude of these constraint values.

Fundamentally, scalarization combines objectives in reward space, whereas our approach combines objectives in \emph{distribution} space, thus making it invariant to the scale of rewards. In principle, our approach can be combined with any RL algorithm, regardless of whether it is off-policy or on-policy. We combine it with maximum a posteriori policy optimization (MPO) \cite{abdolmaleki2018relative,Abdolmaleki_2018}, an off-policy actor-critic RL algorithm, and V-MPO \cite{Song2020V-MPO}, an on-policy variant of MPO. We call these two algorithms multi-objective MPO (MO-MPO) and multi-objective V-MPO (MO-V-MPO), respectively.

Our main contribution is providing a distributional view on MORL, which enables scale-invariant encoding of preferences. We show that this is a theoretically-grounded approach, that arises from taking an RL-as-inference perspective of MORL. Empirically, we analyze the mechanics of MO-MPO and show it finds all Pareto-optimal policies in a popular MORL benchmark task. Finally, we demonstrate that MO-MPO and MO-V-MPO outperform scalarized approaches on multi-objective tasks across several challenging high-dimensional continuous control domains (\figref{fig:domains}).
\section{Related Work}

\subsection{Multi-Objective Reinforcement Learning}
\label{sec:relatedwork}
Multi-objective reinforcement learning (MORL) algorithms are either \emph{single-policy} or \emph{multiple-policy} \cite{Vamplew_2011}. Single-policy approaches seek to find the optimal policy for a given scalarization of the multi-objective problem. Often this scalarization is linear, but other choices have also been explored \cite{Moffaert_2013}.

However, the scalarization may be unknown at training time, or it may change over time. \emph{Multiple-policy} approaches handle this by finding a set of policies that approximates the true Pareto front.
Some approaches repeatedly call a single-policy MORL algorithm with strategically-chosen scalarizations \cite{Natarajan_2005,Roijers_2014,Mossalem_2016,Zuluaga_2016}. Other approaches learn a set of policies simultaneously, by using a multi-objective variant of the Q-learning update rule \cite{Barrett_2008,Moffaert_2014,Reymond_2019,Yang_2019} 
or by modifying gradient-based policy search \cite{Parisi_2014,Pirotta_2015}.

Most existing approaches for finding the Pareto front are limited to discrete state and action spaces, in which tabular algorithms are sufficient. Although recent work combining MORL with deep RL handles high-dimensional observations, this is in domains with low-dimensional and usually discrete action spaces \citep{Mossalem_2016,vanSeijen_2017,Friedman_2018,Abels_2019,Reymond_2019,Yang_2019,Nottingham_2019}. In contrast, we evaluate our approach on continuous control tasks with more than 20 action dimensions.\footnote{MORL for continuous control tasks is difficult because the policy can no longer output arbitrary action distributions, which limits how well it can compromise between competing objectives. In state-of-the-art RL algorithms for continuous control, policies typically output a single action (e.g., D4PG \citep{BarthMaron_2018}) or a Gaussian (e.g., PPO \citep{Schulman_2017}, MPO \citep{Abdolmaleki_2018}, and SAC \citep{Haarnoja_2018}).}

A couple of recent works have applied deep MORL to find the Pareto front in continuous control tasks; these works assume scalarization and rely on additionally learning either a meta-policy \citep{Xi_2019} or inter-objective relationships \citep{Zhang_2019}. We take an orthogonal approach to existing approaches: one encodes preferences via constraints on the influence of each objective on the policy update, instead of via scalarization.
MO-MPO can be run multiple times, with different constraint settings, to find a Pareto front of policies.

\subsection{Constrained Reinforcement Learning}
An alternate way of setting preferences is to enforce that policies meet certain constraints. For instance, threshold lexicographic ordering approaches optimize a (single) objective while meeting specified threshold values on the other objectives \cite{Gabor_1998}, optionally with slack \cite{Wray_2015}. Similarly, safe RL is concerned with learning policies that optimize a scalar reward while not violating safety constraints \cite{Achiam_2017,Chow_2018}; this has also been studied in the off-policy batch RL setting \cite{Le_2019}. Related work minimizes costs while ensuring the policy meets a constraint on the minimum expected return \cite{Bohez_2019}, but this requires that the desired or achievable reward is known a priori.
In contrast, MO-MPO does not
require knowledge of the scale of rewards. In fact, often there is no easy way to specify constraints on objectives, e.g., it is difficult to figure out a priori how much actuator effort a robot will need to use to perform a task.

\subsection{Multi-Task Reinforcement Learning}
Multi-task reinforcement learning can also be cast as a MORL problem. Generally these algorithms learn a separate policy for each task, with shared learning across tasks \citep{teh2017distral,riedmiller2018learning,wulfmeier2019regularized}. In particular, Distral \citep{teh2017distral} learns a shared prior that regularizes the per-task policies to be similar to each other, and thus captures essential structure that is shared across tasks. MO-MPO differs in that the goal is to learn a \emph{single} policy that must \emph{trade off} across different objectives.

Other multi-task RL algorithms seek to train a single agent to solve different tasks, and thus need to handle different reward scales across tasks. Prior work uses adaptive normalization for the targets in value-based RL, so that the agent cares equally about all tasks \cite{van2016learning,hessel2019multi}. Similarly, prior work in multi-objective optimization has dealt with objectives of different units and/or scales by normalizing objectives to have similar magnitudes \citep{Marler_2005,Grodzevich_2006,Kim_2006,Daneshmand_2017,Ishibuchi_2017}. MO-MPO can also be seen as doing adaptive normalization, but for \emph{any} preference over objectives, not just equal preferences.

In general, invariance to reparameterization of the function approximator has been investigated in optimization literature resulting in, for example, natural gradient methods \citep{Martens14}. The common tool here is measuring distances in function space instead of parameter space, using KL-divergence. Similarly in this work, to achieve invariance to the scale of objectives, we use KL-divergence over policies to encode preferences.

\section{Background and Notation}
\label{sec:background}

\prg{Multi Objective Markov Decision Process} In this paper, we consider a multi-objective RL problem defined by a multi-objective Markov Decision Process (MO-MDP). The MO-MDP consists of states $s \in \mathcal{S}$ and actions $a \in \mathcal{A}$, an initial state distribution $p(s_0)$, transition probabilities $p(s_{t+1} | s_t, a_t)$ which define the probability of changing from state $s_t$ to $s_{t+1}$ when taking action $a_{t}$, reward functions $\{r_k(s, a) \in \mathbb{R}\}_{k=1}^N$ per objective $k$, and a discount factor $\gamma \in [0, 1)$. We define our policy $\pi_\theta(a | s)$ as a state conditional distribution over actions parametrized by $\theta$. Together with the transition probabilities, this gives rise to a state visitation distribution $\mu(s)$.
We also consider per-objective action-value functions. The action-value function for objective $k$ is defined as the expected return (i.e., cumulative discounted reward) from choosing action $a$ in state $s$ for objective $k$ and then following policy $\pi$: $Q_k^\pi(s,a) = \mathbb{E}_\pi \lbrack \sum_{t=0}^\infty \gamma^t r_k(s_t, a_t) | s_0=s, a_0 = a]$. We can represent this function using the recursive expression $Q_k^\pi(s_t, a_t) = \mathbb{E}_{p(s_{t+1} | s_t, a_t)} \big[ r_k(s_t, a_t) + \gamma V_k^\pi(s_{t+1}) \big]$, where $V_k^\pi(s) = \mathbb{E}_\pi[ Q_k^\pi(s,a) ]$ is the value function of $\pi$ for objective $k$.

\prg{Problem Statement} For any MO-MDP there is a set of nondominated policies, i.e., the Pareto front. 
A policy is nondominated if there is no other policy that improves its expected return for an objective without reducing the expected return of at least one other objective.
Given a preference setting, our goal is to find a nondominated policy $\pi_{\theta}$ that satisfies those preferences. In our approach, a setting of constraints does not directly correspond to a particular scalarization, but we show that by varying these constraint settings, we can indeed trace out a Pareto front of policies.

\section{Method}

We propose a policy iteration algorithm for multi-objective RL. Policy iteration algorithms decompose the RL problem into two sub-problems and iterate until convergence:
\begin{enumerate}[noitemsep,topsep=0pt]
    \item \emph{Policy evaluation}: estimate Q-functions given policy
    \item \emph{Policy improvement}: update policy given Q-functions
\end{enumerate}
Algorithm \ref{Alg:MO-MPO} summarizes this two-step multi-objective policy improvement procedure. In Appendix E, we explain how this can be derived from the ``RL as inference'' perspective.

We describe multi-objective MPO in this section and explain multi-objective V-MPO in Appendix D. When there is only one objective, MO-(V-)MPO reduces to (V-)MPO.

\begin{algorithm}[t]
\small
\caption{MO-MPO: One policy improvement step}\label{Alg:MO-MPO}
\begin{algorithmic}[1]
\STATE {\bf given} batch size ($L$), number of actions to sample ($M$), ($N$) Q-functions $\{Q_k^{\pi_{\textrm{old}}}(s,a)\}_{k=1}^N$, preferences $\{\epsilon_k\}_{k=1}^N$, previous policy $\pi_\textrm{old}$, previous temperatures $\{\eta_k\}_{k=1}^N$, replay buffer $\mathcal{D}$, first-order gradient-based optimizer $\mathcal{O}$
\STATE
\STATE {\bf initialize} $\pi_\theta$ from the parameters of $\pi_\textrm{old}$

\REPEAT
\STATE {\textbf{// Collect dataset} $\{s^i, a^{ij}, Q_k^{ij}\}_{i,j,k}^{L,M,N}$\textbf{, where}}
\STATE {\textbf{//} $M$ \textbf{actions} $a^{ij} \sim \pi_\textrm{old}(a|s^i)$ \textbf{and} $Q_k^{ij} = Q_k^{\pi_{\textrm{old}}}(s^i, a^{ij})$}
\STATE
\STATE {\bf // Compute action distribution for each objective}
\FOR{k = 1, \dots, $N$}
\STATE \scalebox{0.95}[1]{$
\delta_{\eta_k} \leftarrow \nabla_{\eta_k} \eta_k\epsilon_k+\eta_k\sum_{i}^L\frac{1}{L}\log\left(\sum_j^M\frac{1}{M}\exp\Big(\frac{Q_k^{ij}}{\eta_k}\Big)\right)
$}
\STATE {Update $\eta_k$ based on $\delta_{\eta_k}$, using optimizer $\mathcal{O}$}
\STATE $q_k^{ij} \propto \exp(\frac{Q_k^{ij}}{\eta_k})$
\ENDFOR
\STATE
\STATE {\bf // Update parametric policy}
\STATE $\delta_\pi \leftarrow -\nabla_\theta \sum_i^L \sum_j^M \sum_k^N q_k^{ij} \log \pi_{\theta}(a^{ij}|s^i)$
\STATE {\hspace{0.8em} (subject to additional KL regularization, see \sref{sec:supervisedstep})}

\STATE {Update $\pi_\theta$ based on $\delta_\pi$, using optimizer $\mathcal{O}$}
\STATE
\UNTIL{fixed number of steps}
\STATE return $\pi_\textrm{old}=\pi_\theta$
\end{algorithmic}
\end{algorithm}

\subsection{Multi-Objective Policy Evaluation}
\label{sec:policyevalstep}
In this step we learn Q-functions to evaluate the previous policy $\pi_\textrm{old}$. We train a separate Q-function per objective, following the Q-decomposition approach \citep{Russell_2003}. In principle, any Q-learning algorithm can be used, as long as the target Q-value is computed with respect to $\pi_\textrm{old}$.\footnote{\citet{Russell_2003} prove critics suffer from ``illusion of control'' if they are trained with conventional Q-learning \citep{Watkins_1989}. In other words, if each critic computes its target Q-value based on its own best action for the next state, then they overestimate Q-values, because in reality the parametric policy $\pi_\theta$ (that considers all critics' opinions) is in charge of choosing actions.}
In this paper, we use the Retrace objective \citep{Munos_2016} to learn a Q-function $Q^{\pi_\textrm{old}}_k(s, a; \phi_k)$ for each objective $k$, parameterized by $\phi_k$, as follows:
\begin{align*}
\min_{{\{\phi_k\}}_1^N} \sum_{k=1}^N \mathbb{E}_{(s,a)\sim \mathcal{D}} \Big[ \big( \hat{Q}_k^{\textrm{ret}}(s, a) - Q^{\pi_\textrm{old}}_k(s, a;\phi_k))^2 \Big],
\end{align*}
where $\hat{Q}_k^{\textrm{ret}}$ is the Retrace target for objective $k$ and the previous policy $\pi_\textrm{old}$, and $\mathcal{D}$ is a replay buffer containing gathered transitions. See Appendix C for details.

\subsection{Multi-Objective Policy Improvement}
Given the previous policy $\pi_\textrm{old}(a|s)$ and associated Q-functions $\{Q_k^{\pi_\textrm{old}}(s,a)\}_{k=1}^{N}$, our goal is to improve the previous policy for a given visitation distribution $\mu(s)$.\footnote{In practice,
we use draws from the replay buffer to estimate expectations over the visitation distribution $\mu(s)$.}
To this end, we learn an action distribution for each Q-function and combine these to obtain the next policy $\pi_\textrm{new}(a|s)$. 
This is a multi-objective variant of the two-step policy improvement procedure employed by MPO \cite{Abdolmaleki_2018}.

\textbf{In the first step}, for each objective $k$ we learn an improved action distribution $q_k(a|s)$ such that $\mathbb{E}_{q_k(a|s)}[ Q^{\pi_\textrm{old}}_k(s,a)] \geq \mathbb{E}_{\pi_\textrm{old}(a|s)}[ Q^{\pi_\textrm{old}}_k(s,a)]$, where states $s$ are drawn from a visitation distribution $\mu(s)$.

\textbf{In the second step}, we combine and distill the improved distributions $q_k$ into a new parametric policy $\pi_\textrm{new}$ (with parameters $\theta_\textrm{new}$) by minimizing the KL-divergence between the distributions and the new parametric policy, i.e,
\begin{equation}
    \theta_\textrm{new} = \argmin_{\theta} \sum_{k=1}^N \mathbb{E}_{\mu(s)}\Big[\mathrm{KL}\Big(q_k(a | s) \| \pi_\theta(a | s)\Big)\Big] \, .
\label{eq:klmin}
\end{equation}
This is a supervised learning loss that performs maximum likelihood estimate of distributions $q_k$. 
Next, we will explain these two steps in more detail.
\subsubsection{Obtaining action distributions per objective (Step 1)}
To obtain the per-objective improved action distributions $q_k(a|s)$, we optimize the standard RL objective for {\it each objective} $Q_{k}$:
\begin{align}
\label{eq:qk_opt}
&\max_{q_{k}} \int_s \mu(s) \int_a q_k(a|s) \, Q_{k}(s,a) \diff a \diff s\\
&\textrm{s.t.} \int_s \mu(s) \, \textrm{KL}(q_k(a|s) \| \pi_\textrm{old}(a|s)) \diff s < \epsilon_k \, , \nonumber
\end{align}
where $\epsilon_k$ denotes the allowed expected KL divergence for objective $k$. We use these $\epsilon_k$ to encode preferences over objectives. More concretely, $\epsilon_k$ defines the allowed influence of objective $k$ on the change of the policy.

For nonparametric action distributions $q_k(a|s)$, we can solve this constrained optimization problem in closed form for each state $s$ sampled from $\mu(s)$ \cite{Abdolmaleki_2018},
\begin{equation}
    q_k(a|s) \propto \pi_\textrm{old}(a|s) \exp\Big(\frac{Q_k(s,a)}{\eta_k}\Big) \, ,
\end{equation}
where the temperature $\eta_k$ is computed based on the corresponding $\epsilon_k$, by solving the following convex dual function:
\begin{align}
\label{eq:eta}
\eta_k = &\argmin_{\eta}
\eta \, \epsilon_k \, + \\ &\eta \int_s \mu(s) \log \int_a \pi_\textrm{old}(a|s)\exp\Big(\frac{Q_k(s,a)}{\eta}\Big) \diff a \diff s \, . \nonumber
\end{align}
In order to evaluate $q_k(a|s)$ and the integrals in \eref{eq:eta}, we draw $L$ states from the replay buffer and, for each state, sample $M$ actions from the current policy $\pi_\textrm{old}$.
In practice, we maintain one temperature parameter $\eta_k$ per objective. We found that optimizing the dual function by performing a few steps of gradient descent on $\eta_k$ is effective, and we initialize with the solution found in the previous policy iteration step. Since $\eta_k$ should be positive, we use a projection operator after each gradient step to maintain $\eta_k > 0$.
Please refer to Appendix C for derivation details.

\prg{Application to Other Deep RL Algorithms} Since the constraints $\epsilon_k$ in \eref{eq:qk_opt} encode the preferences over objectives, solving this optimization problem with good satisfaction of constraints is key for learning a policy that satisfies the desired preferences. For nonparametric action distributions $q_k(a|s)$, we can satisfy these constraints exactly. One could use any policy gradient method \citep[e.g.][]{schulman15,Schulman_2017,svg,Haarnoja_2018} to obtain $q_k(a|s)$ in a \emph{parametric} form instead. However, solving the constrained optimization for parametric $q_k(a|s)$ is not exact, and the constraints may not be well satisfied, which impedes the use of $\epsilon_k$ to encode preferences.
Moreover, assuming a parametric $q_k(a|s)$ requires maintaining a function approximator (e.g., a neural network) per objective, which can significantly increase the complexity of the algorithm and limits scalability.

\prg{Choosing $\epsilon_k$} It is more intuitive to encode preferences via $\epsilon_k$ rather than via scalarization weights, because the former is invariant to the scale of rewards. In other words, having a desired preference across objectives narrows down the range of reasonable choices for $\epsilon_k$, but does not narrow down the range of reasonable choices for scalarization weights. In order to identify reasonable scalarization weights, a RL practitioner needs to additionally be familiar with the scale of rewards for each objective. In practice, we have found that learning performance is robust to a wide range of scales for $\epsilon_k$. It is the \emph{relative} scales of the $\epsilon_k$ that matter for encoding preferences over objectives---the larger a particular $\epsilon_k$ is with respect to others, the more that objective $k$ is preferred. On the other hand, if $\epsilon_{k} = 0$, then objective $k$ will have no influence and will effectively be ignored. In Appendix A.1, we provide suggestions for setting $\epsilon_k$, given a desired preference over objectives.

\subsubsection{Fitting a new parametric policy (Step 2)}
\label{sec:supervisedstep}
In the previous section, for each objective $k$, we have obtained an improved action distribution $q_k(a|s)$. Next, we want to combine these distributions to obtain a single parametric policy that trades off the objectives according to the constraints $\epsilon_k$ that we set. For this, we solve a supervised learning problem that fits a parametric policy to the per-objective action distributions from step 1,
\begin{align}
\theta_\textrm{new} = & \argmax_{\theta}  \sum_{k=1}^N \int_s \mu(s) \int_a q_k(a|s)\log \pi_\theta(a|s)\diff a\diff s  \nonumber \\ 
&\textrm{s.t.} \int_s \mu(s) \, \KL(\pi_\textrm{old}(a|s) \,\|\, \pi_\theta(a|s) ) \diff s < \beta \, ,
\label{eq:klpolicy}
\end{align}
where $\theta$ are the parameters of our policy (a neural network) and the KL constraint enforces a trust region of size $\beta$ that limits the overall change in the parametric policy. The KL constraint in this step has a regularization effect that prevents the policy from overfitting to the sample-based action distributions, and therefore avoids premature convergence and improves stability of learning \cite{schulman15,abdolmaleki2018relative,Abdolmaleki_2018}.

Similar to the first policy improvement step, we evaluate the integrals by using the $L$ states sampled from the replay buffer and the $M$ actions per state sampled from the old policy. In order to optimize \eref{eq:klpolicy} using gradient descent, we employ Lagrangian relaxation, similar to in MPO \citep{abdolmaleki2018relative} (see Appendix C for more detail).

\section{Experiments: Toy Domains}
In the empirical evaluation that follows, we will first demonstrate the mechanics and scale-invariance of MO-MPO in a single-state environment (\sref{sec:mab}), and then show that MO-MPO can find all Pareto-dominant policies in a popular MORL benchmark (\sref{sec:dst}). Finally, we show the benefit of using MO-MPO in high-dimensional continuous control domains, including on a real robot (\sref{sec:continuouscontrol}). Appendices A and B contain a detailed description of all domains and tasks, experimental setups, and implementation details.

\prg{Baselines} The goal of our empirical evaluation is to analyze the benefit of using our proposed multi-objective policy improvement step (\sref{sec:supervisedstep}), that encodes preferences over objectives via constraints $\epsilon_k$ on expected KL-divergences, rather than via weights $w_k$. Thus, we primarily compare MO-MPO against \emph{scalarized MPO}, which relies on linear scalarization weights $w_k$ to encode preferences. The only difference between MO-MPO and scalarized MPO is the policy improvement step: for scalarized MPO, a single improved action distribution $q(a|s)$ is computed, based on $\sum_k w_k Q_k(s,a)$ and a single KL constraint $\epsilon$.

State-of-the-art approaches that combine MORL with deep RL assume linear scalarization as well, either  learning a separate policy for each setting of weights \citep{Mossalem_2016} or learning a single policy conditioned on scalarization weights \citep{Friedman_2018, Abels_2019}. Scalarized MPO addresses the former problem, which is easier. The policy evaluation step in scalarized MPO is analagous to \emph{scalarized Q-learning}, proposed by \citet{Mossalem_2016}. As we show later in \sref{sec:continuouscontrol}, even learning an optimal policy for a single scalarization is difficult in high-dimensional continuous control domains.

\subsection{Simple World}
\label{sec:mab}
\begin{figure}[t!]
    \centering
    \includegraphics[width=0.5\linewidth]{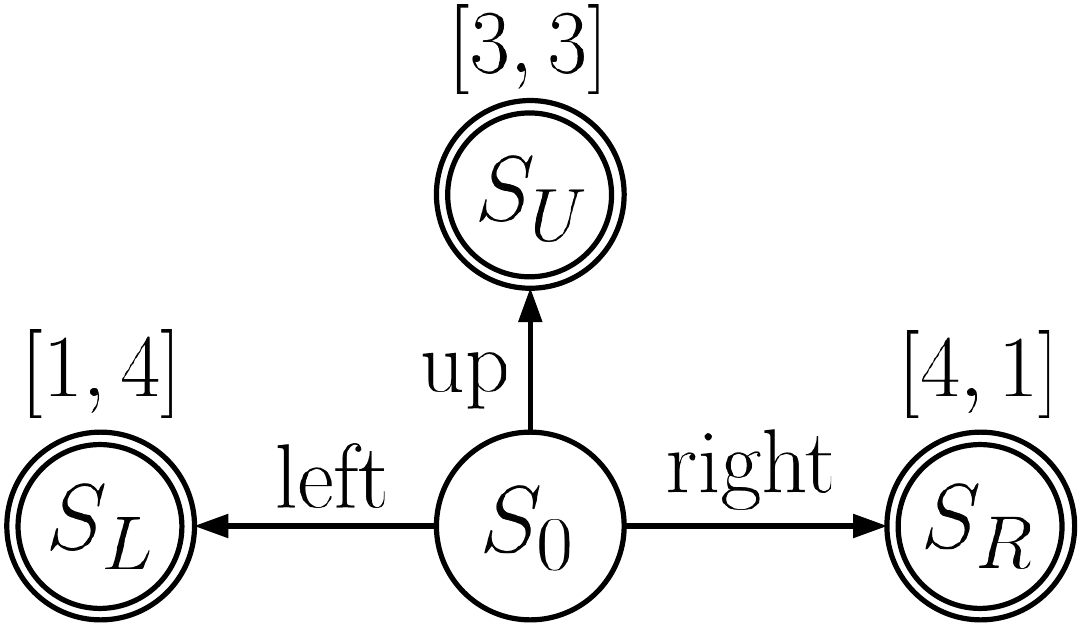}
    \caption{A simple environment in which the agent starts at state $S_0$, and chooses to navigate to one of three terminal states. There are two objectives. Taking the \texttt{left} action, for instance, leads to a reward of $1$ for the first objective and $4$ for the second.}
    \label{fig:simpleworld}
\end{figure}
First, we will examine the behavior of MO-MPO in a simple multi-armed bandit with three actions (\texttt{up}, \texttt{right}, and \texttt{left}) (\figref{fig:simpleworld}), inspired by Russell \& Zimdars \yrcite{Russell_2003}.
We train policies with scalarized MPO and with MO-MPO. The policy evaluation step is exact because the $Q$-value function for each objective is known: it is equal to the reward received for that objective after taking each action, as labeled in \figref{fig:simpleworld}.\footnote{The policy improvement step can also be computed exactly, because solving for the optimal temperature $\eta$ (or $\eta_1$ and $\eta_2$ in the MO-MPO case) is a convex optimization problem, and the KL-constrained policy update is also a convex optimization problem when there is only one possible state. We use CVXOPT \cite{CVXOPT} as our convex optimization solver.}

\begin{figure}[t!]
    \centering
    \includegraphics[width=\linewidth]{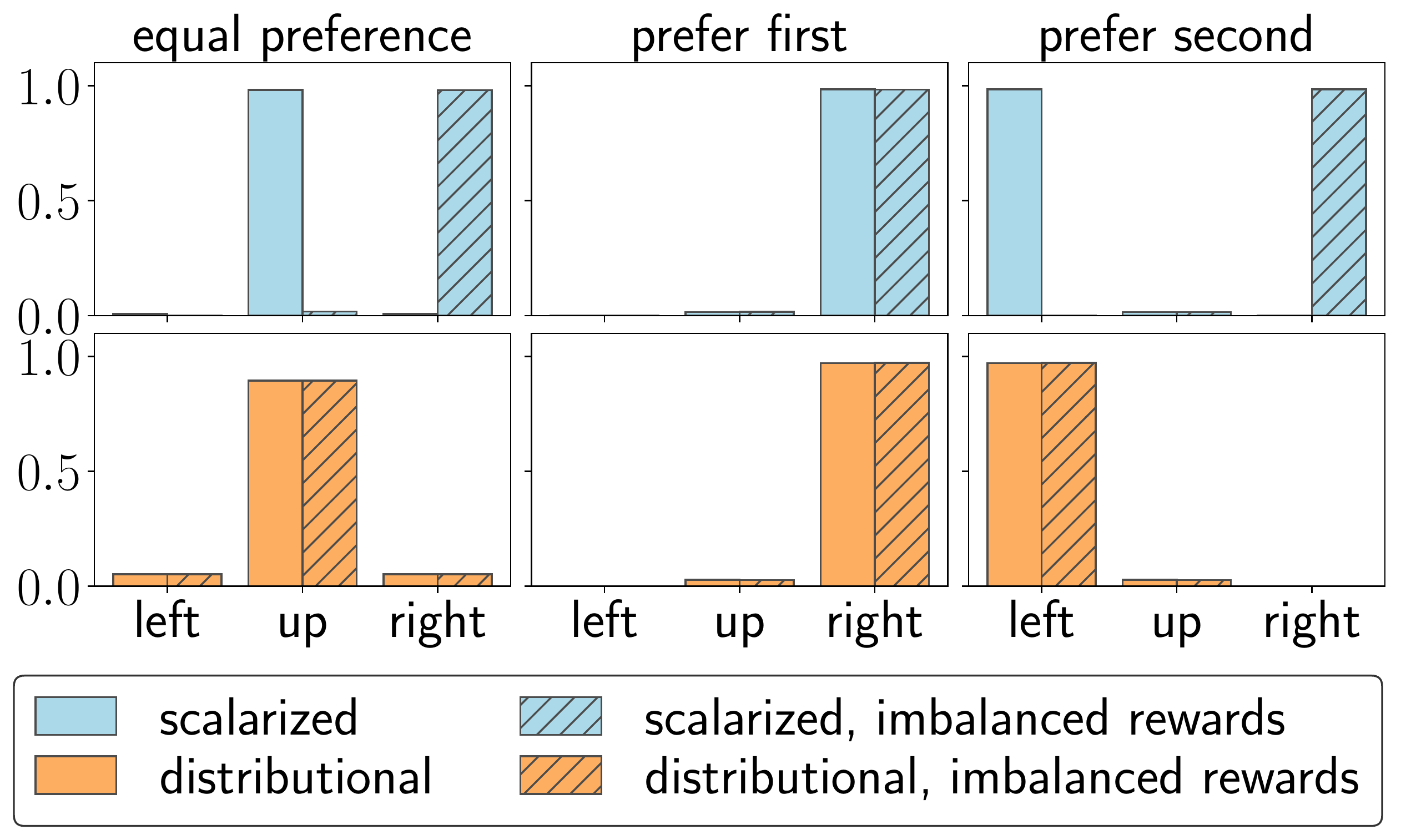}
    \caption{When two objectives have comparable reward scales (solid bars), scalarized MPO (first row) and MO-MPO (second row) learn similar policies, across three different preferences. However, when the scale of the first objective is much higher (striped bars), scalarized MPO shifts to always preferring the first objective. In contrast, MO-MPO is scale-invariant and still learns policies that satisfy the preferences. The y-axis denotes action probability.}
    \label{fig:simple_scale}
\end{figure}

We consider three possible desired preferences: equal preference for the two objectives, preferring the first, and preferring the second. Encoding preferences in scalarized MPO amounts to choosing appropriate linear scalarization weights, and in MO-MPO amounts to choosing appropriate $\epsilon$'s. We use the following weights and $\epsilon$'s:
\begin{itemize}[noitemsep,topsep=0pt]
    \item \emph{equal preference}: weights $[0.5, 0.5]$ or $\epsilon$'s $[0.01, 0.01]$
    \item \emph{prefer first}: weights $[0.9, 0.1]$ or $\epsilon$'s $[0.01, 0.002]$
    \item \emph{prefer second}: weights $[0.1, 0.9]$ or $\epsilon$'s $[0.002, 0.01]$
\end{itemize}
We set $\epsilon = 0.01$ for scalarized MPO. If we start with a uniform policy and run MPO with $\beta = 0.001$ until the policy converges, scalarized MPO and MO-MPO result in similar policies (\figref{fig:simple_scale}, solid bars): \texttt{up} for \emph{equal preference}, \texttt{right} for \emph{prefer first}, and \texttt{left} for \emph{prefer second}.

However, if we make the rewards \emph{imbalanced} by multiplying the rewards obtained for the first objective by $20$ (e.g., \texttt{left} now obtains a reward of $[20, 4]$), we see that the policies learned by scalarized MPO shift to preferring the optimal action for the first objective (\texttt{right}) in both the \emph{equal preference} and \emph{prefer second} cases (\figref{fig:simple_scale}, striped bars). In contrast, the final policies for MO-MPO are the same as for balanced rewards, because in each policy improvement step, MO-MPO optimizes for a separate temperature $\eta_k$ that scales each objective's $Q$-value function. This $\eta_k$ is computed based on the corresponding allowed KL-divergence $\epsilon_k$, so when the rewards for any objective $k$ are multiplied by a factor but $\epsilon_k$ remains the same, the computed $\eta_k$ ends up being scaled by that factor as well, neutralizing the effect of the scaling of rewards (see Eq. \eqref{eq:eta}).

Even in this simple environment, we see that MO-MPO's scale-invariant way of encoding preferences is valuable. In more complex domains, in which the $Q$-value functions must be learned in parallel with the policy, the (automatic) dynamic adjustment of temperatures $\eta_k$ per objective becomes more essential (\sref{sec:continuouscontrol}).

\begin{figure}[t!]
    \centering
    \includegraphics[width=\linewidth]{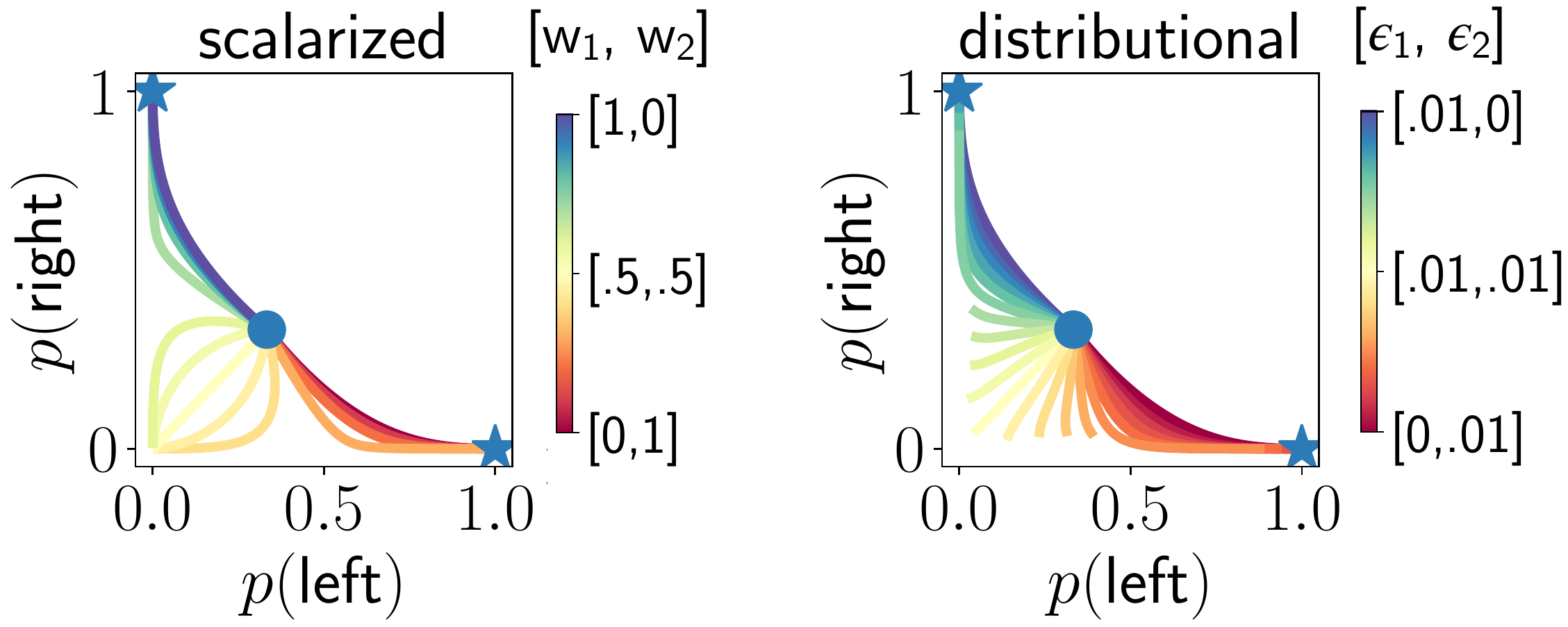}
    \caption{A visualization of policies during learning---each curve corresponds to a particular setting of weights (left) or $\epsilon$'s (right). Policies are initialized as uniform (the blue dot), and are trained until convergence. Each point $(x,y)$ corresponds to the policy with $p($\texttt{left}$) = x$, $p($\texttt{right}$) = y$, and $p($\texttt{up}$) = 1 - x - y$. The top left and bottom right blue stars denote the optimal policy for the first and second objectives, respectively.}
    \label{fig:simple_simplex}
\end{figure}

The scale of $\epsilon_k$ controls the amount that objective $k$ can influence the policy's update.
If we set $\epsilon_1 = 0.01$ and sweep over the range from $0$ to $0.01$ for $\epsilon_2$, the resulting policies go from always picking \texttt{right}, to splitting probability across \texttt{right} and \texttt{up}, to always picking \texttt{up} (\figref{fig:simple_simplex}, right). In contrast, setting weights leads to policies quickly converging to placing all probability on a single action (\figref{fig:simple_simplex}, left). We hypothesize this limits the ability of scalarized MPO to explore and find compromise policies (that perform well with respect to all objectives) in more challenging domains.


\subsection{Deep Sea Treasure}
\label{sec:dst}

\begin{figure}[t!]
    \centering
    \includegraphics[width=1.\linewidth]{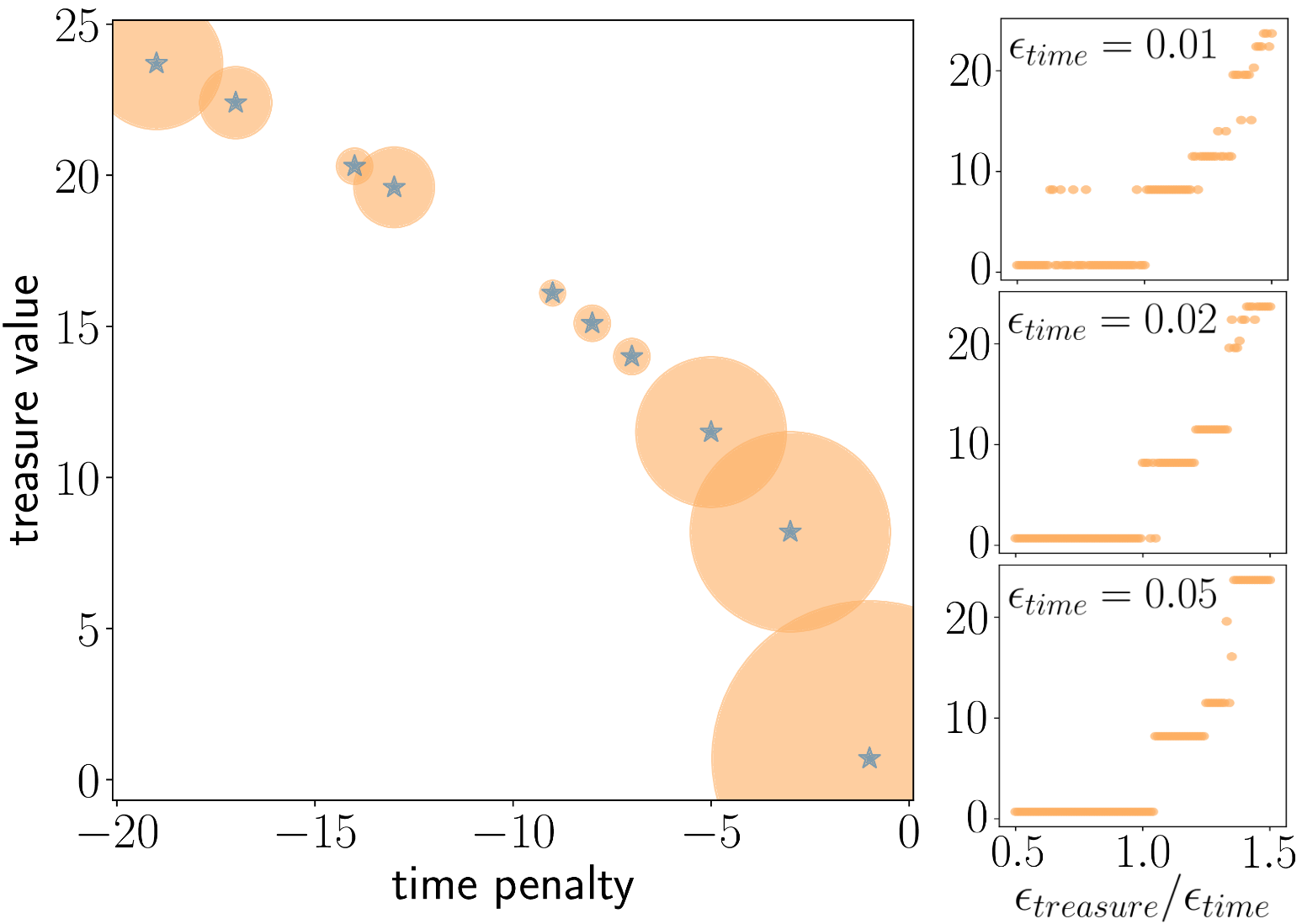}  
    \caption{Left: Blue stars mark the true Pareto front for Deep Sea Treasure. MO-MPO, with a variety of settings for $\epsilon_k$, discovers all points on the true Pareto front. The area of the orange circles is proportional to the number of $\epsilon_k$ settings that converged to that point. Right: As more preference is given to the treasure objective (i.e., as $x$ increases), policies tend to prefer higher-value treasures. Each orange dot in the scatterplot corresponds to a particular setting of $\epsilon_k$.}
    \label{fig:dst}
\end{figure}

An important quality of any MORL approach is the ability to find a variety of policies on the true Pareto front \cite{Roijers_2013}. We demonstrate this in Deep Sea Treasure (DST) \cite{Vamplew_2011}, a popular benchmark for testing MORL approaches. DST consists of a $11 \plh 10$ grid world with ten treasure locations. The agent starts in the upper left corner ($s_0 = (0, 0)$) and has a choice of four actions (moving one square \texttt{up}, \texttt{right}, \texttt{down}, and \texttt{left}). When the agent picks up a treasure, the episode terminates. The agent has two objectives: treasure value and time penalty. The time penalty is $-1$ for each time step that the agent takes, and farther-away treasures have higher treasure values. We use the treasure values in Yang et al. \yrcite{Yang_2019}.

We ran scalarized MPO with weightings $[w, 1-w]$ and $w \in [0, 0.01, 0.02, \dots, 1]$. All policies converged to a point on the true Pareto front, and all but three found the optimal policy for that weighting. In terms of coverage, policies were found for eight out of ten points on the Pareto front.\footnote{Since scalarized MPO consistently finds the optimal policy for any given weighting in this task, with a more strategic selection of weights, we expect policies for all ten points would be found.}

We ran MO-MPO on this task as well, for a range of $\epsilon$: $\epsilon_\text{time} \in [0.01, 0.02, 0.05]$ and $\epsilon_\text{treasure} = c * \epsilon_\text{time}$, where $c \in [0.5, 0.51, 0.52, \dots, 1.5]$. All runs converged to a policy on the true Pareto front, and MO-MPO finds policies for all ten points on the front (\figref{fig:dst}, left). Note that it is the \emph{ratio} of $\epsilon$'s that matters, rather than the exact settings---across all settings of $\epsilon_\text{time}$, similar ratios of $\epsilon_\text{treasure}$ to $\epsilon_\text{time}$ result in similar policies; as this ratio increases, policies tend to prefer higher-value treasures (\figref{fig:dst}, right).
\section{Experiments: Continuous Control Domains}
\label{sec:continuouscontrol}
The advantage of encoding preferences via $\epsilon$'s, rather than via weights, is apparent in more complex domains. We compared our approaches, MO-MPO and MO-V-MPO, against scalarized MPO and V-MPO in four high-dimensional continuous control domains, in MuJoCo \cite{Todorov_2012} and on a real robot. The domains we consider are:

\textbf{Humanoid:} We use the humanoid \textbf{\emph{run}} task defined in DeepMind Control Suite \cite{Tassa_2018}. Policies must optimize for horizontal speed $h$ while minimizing energy usage. The task reward is $\min(h/10, 1)$ where $h$ is in meters per second, and the energy usage penalty is action $\ell2$-norm. The humanoid has 21 degrees of freedom, and the observation consists of joint angles, joint velocities, head height, hand and feet positions, torso vertical orientation, and center-of-mass velocity, for a total of 67 dimensions.

\textbf{Shadow Hand:} We consider three tasks on the Shadow Dexterous Hand: \textbf{\emph{touch}}, \textbf{\emph{turn}}, and \textbf{\emph{orient}}. In the \textbf{\emph{touch}} and \textbf{\emph{turn}} tasks, policies must complete the task while minimizing ``pain.'' A sparse task reward of $1.0$ is given for pressing the block with greater than $5$N of force or for turning the dial from a random initial location to the target location. The pain penalty penalizes the robot for colliding with objects at high speed; this penalty is defined as in \citet{Huang_2019}. In the \textbf{\emph{orient}} task, there are three aligned objectives: touching the rectangular peg, lifting it to a given height, and orienting it to be perpendicular to the ground. All three rewards are between 0 and 1. The Shadow Hand has five fingers and 24 degrees of freedom, actuated by 20 motors. The observation consists of joint angles, joint velocities, and touch sensors, for a total of 63 dimensions. The \textbf{\emph{touch}} and \textbf{\emph{turn}} tasks terminate when the goal is reached or after 5 seconds, and the \textbf{\emph{orient}} task terminates after 10 seconds.

\textbf{Humanoid Mocap:} We consider a large-scale humanoid motion capture tracking task, similar to in \citet{Peng_2018}, in which policies must learn to follow motion capture reference data.\footnote{This task was developed concurrently by \citet{anonymous2020tracking} and does not constitute a contribution of this paper.} There are five objectives, each capturing a different aspect of the similarity of the pose between the simulated humanoid and the mocap target: joint orientations, joint velocities, hand and feet positions, center-of-mass positions, and certain body positions and joint angles. These objectives are described in detail in Appendix B.4. In order to balance these multiple objectives, prior work relied on heavily-tuned reward functions \citep[e.g.][]{Peng_2018}. The humanoid has 56 degrees of freedom and the observation is 1021-dimensional, consisting of proprioceptive observations as well as six steps of motion capture reference frames. In total, we use about 40 minutes of locomotion mocap data, making this an extremely challenging domain.

\textbf{Sawyer Peg-in-Hole:} We train a Rethink Robotics Sawyer robot arm to insert a cylindrical peg into a hole, while minimizing wrist forces. The task reward is shaped toward positioning the peg directly above the hole and increases for insertion, and the penalty is the $\ell1$-norm of Cartesian forces measured by the wrist force-torque sensor. The latter implicitly penalizes contacts and impacts, as well as excessive directional change (due to the gripper's inertia inducing forces when accelerating). We impose a force threshold to protect the hardware---if this threshold is exceeded, the episode is terminated. The action space is the end effector's Cartesian velocity, and the observation is 102-dimensional, consisting of Cartesian position, joint position and velocity, wrist force-torque, and joint action, for three timesteps.

\subsection{Evaluation Metric}
We run MO-(V-)MPO and scalarized (V-)MPO with a wide range of constraint settings $\epsilon_k$ and scalarization weights $w_k$, respectively, corresponding to a wide range of possible desired preferences. (The exact settings are provided in Appendix A.) For tasks with two objectives, we plot the Pareto front found by each approach.
We also compute the \emph{hypervolume} of each found Pareto front; this metric is commonly-used for evaluating MORL algorithms \cite{Vamplew_2011}. Given a set of policies $\Pi$ and a reference policy $r$ that is dominated by all policies in this set, this metric is the hypervolume of the space of all policies that dominate $r$ and are dominated by at least one policy in $\Pi$. We use DEAP \cite{Fortin_2012} to compute hypervolumes.

\subsection{Results: Humanoid and Shadow Hand}
\begin{figure}[t!]
    \centering
    \includegraphics[width=\linewidth]{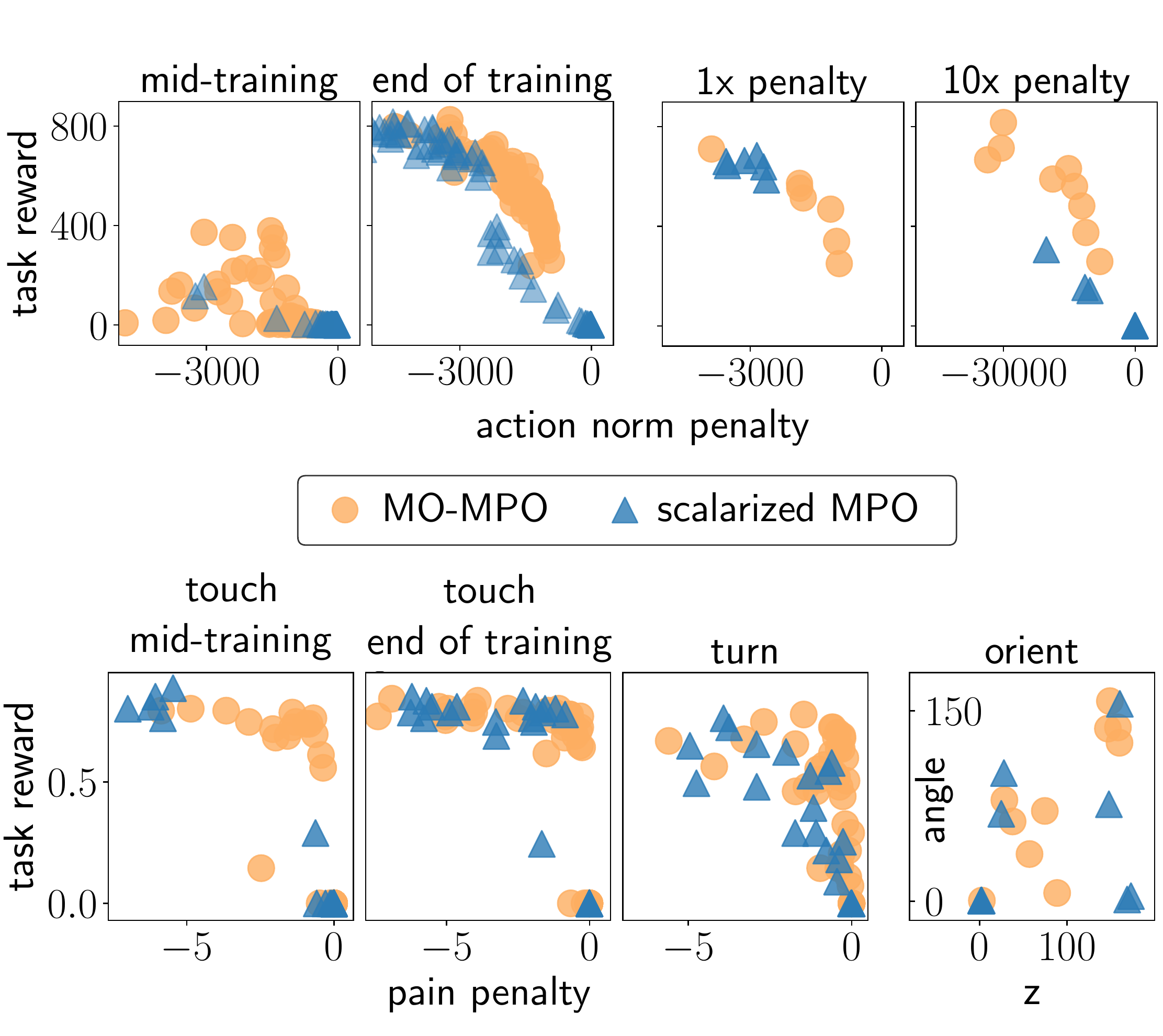}
    \caption{Pareto fronts found by MO-MPO and scalarized MO-MPO for humanoid \textbf{\emph{run}} (top row) and Shadow Hand tasks. Each dot represents a single trained policy. Corresponding hypervolumes are in \tableref{tab:hypervolume}. Task reward is discounted for \textbf{\emph{touch}} and \textbf{\emph{turn}}, with a discount factor of $0.99$. For \textbf{\emph{orient}}, the $x$ and $y$ axes are the total reward for the lift and orientation objectives, respectively.}
    \label{fig:pareto_curves}
    \vspace{-1.5em}
\end{figure}

In the \textbf{\emph{run}}, \textbf{\emph{touch}}, and \textbf{\emph{turn}} tasks, the two objectives are competing---a very high preference for minimizing action norm or pain, as opposed to getting task reward, will result in a policy that always chooses zero-valued actions. Across these three tasks, the Pareto front found by MO-MPO is superior to the one found by scalarized MPO, with respect to the hypervolume metric (\tableref{tab:hypervolume}).\footnote{We use hypervolume reference points of $[0, -10^4]$ for \textbf{\emph{run}}, $[0, -10^5]$ for \textbf{\emph{run}} with $10\plh$ penalty, $[0, -20]$ for \textbf{\emph{touch}} and \textbf{\emph{turn}}, and the zero vector for \textbf{\emph{orient}} and humanoid mocap.} In particular, MO-MPO finds more policies that perform well with respect to \emph{both} objectives, i.e., are in the upper right portion of the Pareto front. MO-MPO also speeds up learning on \textbf{\emph{run}} and \textbf{\emph{touch}} (\figref{fig:pareto_curves}). Qualitatively, MO-MPO trains \textbf{\emph{run}} policies that look more natural and ``human-like''; videos are at \url{http://sites.google.com/view/mo-mpo}.

When we scale the action norm penalty by $10\plh$ for \textbf{\emph{run}}, scalarized MPO policies no longer achieve high task reward, whereas MO-MPO policies do (\figref{fig:pareto_curves}, top right). This supports that MO-MPO's encoding of preferences is indeed scale-invariant.
When the objectives are aligned and have similarly scaled rewards, as in the \textbf{\emph{orient}} task, MO-MPO and scalarized MPO perform similarly, as expected.

\prg{Ablation} We also ran vanilla MPO on humanoid \textbf{\emph{run}}, with the same range of weight settings as for scalarized MPO, to investigate how useful Q-decomposition is. In vanilla MPO, we train a single critic on the scalarized reward function, which is equivalent to removing Q-decomposition from scalarized MPO. Vanilla MPO trains policies that achieve similar task reward (up to 800), but with twice the action norm penalty (up to $-10^4$). As a result, the hypervolume of the Pareto front that vanilla MPO finds is more than a magnitude worse than that of scalarized MPO and MO-MPO ($2.2 \plh 10^6$ versus $2.6 \plh 10^7$ and $6.9 \plh 10^7$, respectively).

\begin{table}[t]
    \small
    \centering
    \setlength\tabcolsep{3pt}
    \begin{tabular}{lcc}
        \toprule \hline
        \textbf{Task} & \textbf{scalarized MPO} & \textbf{MO-MPO} \Tstrut \\
        \midrule
        Humanoid \textbf{\emph{run}}, mid-training          & $1.1 \plh 10^{6}$ & $\boldsymbol{3.3 \plh 10^{6}}$ \\
        Humanoid \textbf{\emph{run}}                        & $6.4 \plh 10^{6}$ & $\boldsymbol{7.1 \plh 10^{6}}$ \\
        \midrule
        Humanoid \textbf{\emph{run}}, $1 \plh$ penalty      & $5.0 \plh 10^{6}$ & $\boldsymbol{5.9 \plh 10^{6}}$ \\
        Humanoid \textbf{\emph{run}}, $10 \plh$ penalty     & $2.6 \plh 10^{7}$ & $\boldsymbol{6.9 \plh 10^{7}}$ \\
        \midrule
        Shadow \textbf{\emph{touch}}, mid-training          & $14.3$  & $\boldsymbol{15.6}$  \\
        Shadow \textbf{\emph{touch}}                        & $16.2$  & $\boldsymbol{16.4}$  \\
        Shadow \textbf{\emph{turn}}                         & $14.4$ & $\boldsymbol{15.4}$ \\
        Shadow \textbf{\emph{orient}}                         & $\boldsymbol{2.8 \plh 10^4}$ & $\boldsymbol{2.8 \plh 10^4}$  \\
        \midrule
        Humanoid Mocap                                      & $\boldsymbol{3.86 \plh 10^{-6}}$ & $3.41 \plh 10^{-6}$ \\
        \midrule
        \bottomrule
    \end{tabular}
    \caption{Hypervolume measurements across tasks and approaches.}
    \label{tab:hypervolume}
    \vspace{-1em}
\end{table}

\subsection{Results: Humanoid Mocap}
The objectives in this task are mostly aligned. In contrast to the other experiments, we use V-MPO \cite{Song2020V-MPO} as the base algorithm because it outperforms MPO in learning this task. In addition, since V-MPO is an on-policy variant of MPO, this enables us to evaluate our approach in the on-policy setting.
Each training run is very computationally expensive, so we train only a handful of policies each.\footnote{For MO-V-MPO, we set all $\epsilon_k = 0.01$. Also, for each objective, we set $\epsilon_k = 0.001$ and set all others to $0.01$. For V-MPO, we fix the weights of four objectives to reasonable values, and try weights of $[0.1, 0.3, 0.6, 1, 5]$ for matching joint velocities.} None of the MO-V-MPO policies are dominated by those found by V-MPO. In fact, although the weights span a wide range of ``preferences'' for the joint velocity, the only policies found by V-MPO that are \emph{not} dominated by a MO-V-MPO policy are those with extreme values for joint velocity reward (either $\leq 0.006$ or $\geq 0.018$), whereas it is between $0.0103$ and $0.0121$ for MO-V-MPO policies.

Although the hypervolume of the Pareto front found by V-MPO is higher than that of MO-V-MPO (\tableref{tab:hypervolume}), finding policies that over- or under- prioritize any objective is undesirable. Qualitatively, the policies trained with MO-V-MPO look more similar to the mocap reference data---they exhibit less feet jittering, compared to those trained with scalarized V-MPO; this can be seen in the corresponding video.

\subsection{Results: Sawyer Peg-in-Hole}
\begin{figure}[t!]
    \centering
    \includegraphics[width=\linewidth]{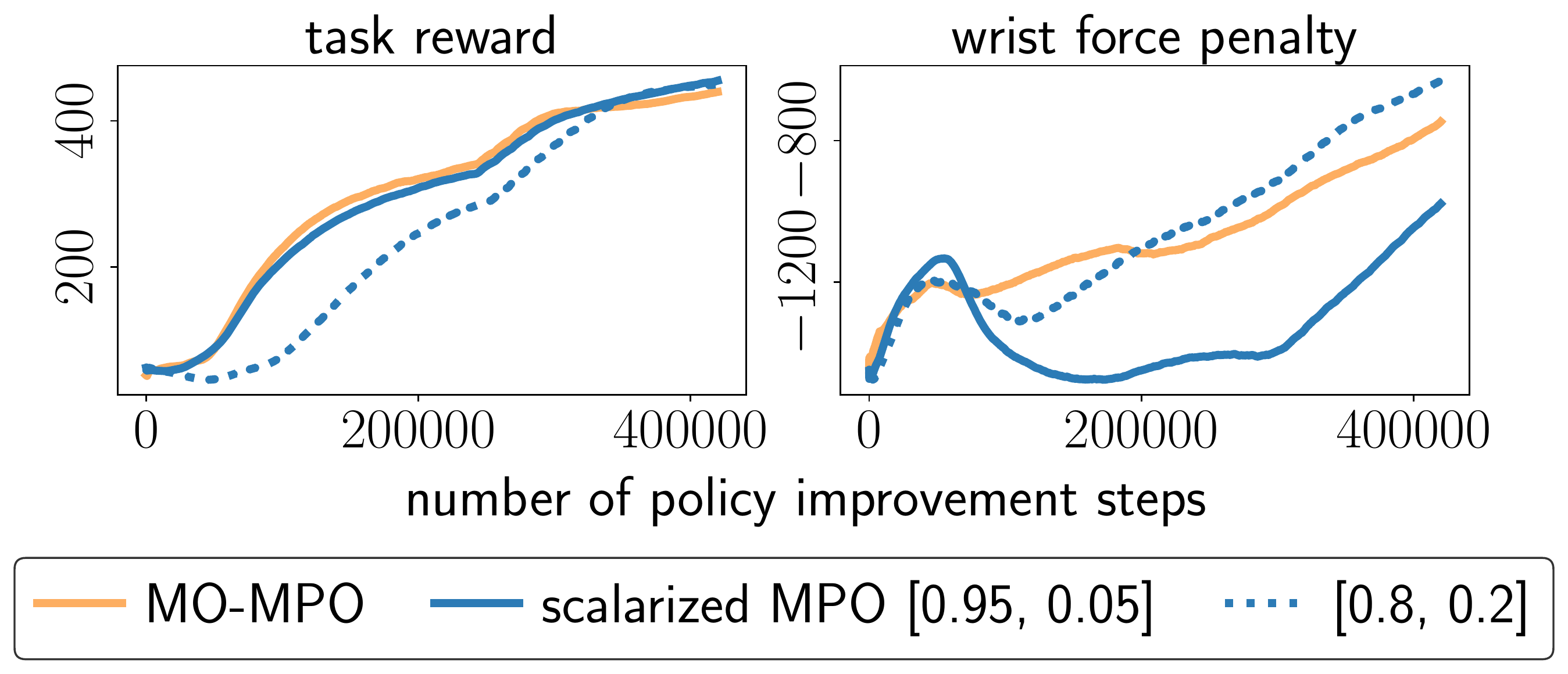}
    \caption{Task reward (left) and wrist force penalty (right) learning curves for the Sawyer peg-in-hole task. The policy trained with MO-MPO quickly learns to optimize \emph{both} objectives, whereas the other two do not. Each line represents a single trained policy.}
    \label{fig:realrobot}
    \vspace{-1em}
\end{figure}

In this task, we would like the robot to prioritize successful task completion, while minimizing wrist forces.  With this in mind, for MO-MPO we set $\epsilon_\text{task} = 0.1$ and $\epsilon_\text{force} = 0.05$, and for scalarized MPO we try $[w_\text{task}, w_\text{force}] = [0.95, 0.05]$ and $[0.8, 0.2]$.
We find that policies trained with scalarized MPO focus all learning on a single objective at the beginning of training; we also observed this in the \textbf{\emph{touch}} task, where scalarized MPO policies quickly learn to either maximize task reward or minimize pain penalty, but not both (\figref{fig:pareto_curves}, bottom left).
In contrast, the policy trained with MO-MPO simultaneously optimizes for \emph{both} objectives throughout training. In fact, throughout training, the MO-MPO policy does just as well with respect to task reward as the scalarized MPO policy that cares more about task reward, and similarly for wrist force penalty (\figref{fig:realrobot}).

\section{Conclusions and Future Work}
In this paper we presented a new distributional perspective on multi objective reinforcement learning, that is derived from the Rl-as-inference perspective. This view leads to two novel multi-objective RL algorithms, namely MO-MPO and MO-V-MPO. We showed these algorithms enable practitioners to encode preferences in a scale-invariant way.
Although MO-(V-)MPO is a single-policy approach to MORL, it is capable of producing a variety of Pareto-dominant policies. A limitation of this work is that we produced this set of policies by iterating over a relatively large number of $\epsilon$'s, which is computationally expensive. In future work, we plan to extend MO-(V-)MPO into a true multiple-policy MORL approach, either by conditioning the policy on settings of $\epsilon$'s or by developing a way to strategically select $\epsilon$'s to train policies for, analogous to what prior work has done for weights (e.g., \citet{Roijers_2014}).
\section*{Acknowledgments}
The authors would like to thank Guillaume Desjardins, Csaba Szepesvari, Jost Tobias Springenberg, Steven Bohez, Philemon Brakel, Brendan Tracey, Jonas Degrave, Jonas Buchli, Leslie Fritz, Chloe Rosenberg, and many others at DeepMind for their support and feedback on this paper.

\bibliography{references}
\bibliographystyle{icml2020}

\clearpage
\appendix

\section{Experiment Details}
In this section, we describe implementation details and specify the hyperparameters used for our algorithm. In all our experiments, the policy and critic(s) are implemented with feed-forward neural networks. We use Adam \citep{Kingma_2015} for optimization.

In our continuous control tasks, the policy returns a Gaussian distribution with a diagonal covariance matrix, i.e, 
$
  \pi_{\a}(\vec a| \vec s) = \mathcal{N}\left(\mu , \vec
  \Sigma \right)
$. The policy is parametrized by a neural network, which outputs the mean $\mu=\mu(\vec s)$ and diagonal Cholesky factors $A=A(\vec s)$, such that $\Sigma = AA^T$. The diagonal factor $A$ has positive diagonal elements enforced by the softplus transform $A_{ii} \leftarrow \log(1 + \exp(A_{ii}))$ to ensure positive definiteness of the diagonal covariance matrix.

\tableref{t:MPO} shows the default hyperparameters that we use for MO-MPO and scalarized MPO, and Table \ref{t:MPO_domainspecific} shows the hyperparameters that differ from these defaults for the humanoid, Shadow Hand, and Sawyer experiments. For all tasks that we train policies with MO-MPO for, a separate critic is trained for each objective, with a shared first layer. The only exception is the action norm penalty in humanoid \textbf{\emph{run}}, which is computed exactly from the action. We found that for both policy and critic networks, layer normalization of the first layer followed by a hyperbolic tangent ($\tanh$) is important for the stability of the algorithm. 

\begin{table}[t]
\small
\centering
\setlength\tabcolsep{3pt}
\begin{tabular}{lc} 
 \toprule \midrule
 \textbf{Hyperparameter} & \textbf{Default} \Tstrut \\
 \midrule
 policy network & \\
 \hspace{2em} layer sizes & $(300, 200)$ \\
 \hspace{2em} take $\tanh$ of action mean? & yes \\
 \hspace{2em} minimum variance & $10^{-12}$ \\
 \hspace{2em} maximum variance & unbounded \\
 \midrule
 critic network(s) & \\
 \hspace{2em} layer sizes & $(400, 400, 300)$ \\
 \hspace{2em} take $\tanh$ of action? & yes \\
 \hspace{2em} Retrace sequence size & $8$ \\
 \hspace{2em} discount factor $\gamma$ & $0.99$ \\
 \midrule
 both policy and critic networks & \\
 \hspace{2em} layer norm on first layer? & yes \\ 
 \hspace{2em} $\tanh$ on output of layer norm? & yes \\
 \hspace{2em} activation (after each hidden layer) & ELU \\
 \midrule
 MPO & \\
 \hspace{2em} actions sampled per state & $20$ \\
 \hspace{2em} default $\epsilon$ & $0.1$ \\
 \hspace{2em} KL-constraint on policy mean, $\beta_{\mu}$ & $10^{-3}$ \\
 \hspace{2em} KL-constraint on policy covariance, $\beta_{\Sigma}$ & $10^{-5}$ \\
 \hspace{2em} initial temperature $\eta$ & $1$ \\
 \midrule
 training & \\
 \hspace{2em} batch size & $512$ \\ 
 \hspace{2em} replay buffer size & $10^{6}$ \\
 \hspace{2em} Adam learning rate & $3 \plh 10^{-4}$ \\ 
 \hspace{2em} Adam $\epsilon$ & $10^{-3}$ \\ 
 \hspace{2em} target network update period & $200$ \\ 
 \midrule
 \bottomrule
\end{tabular}
\caption{Default hyperparameters for MO-MPO and scalarized MPO, with decoupled update on mean and covariance (\sref{sec:policy_fitting}).}
\label{t:MPO}
\end{table}

\begin{table}[t]
\small
\centering
\setlength\tabcolsep{3pt}
\begin{tabular}{lc} 
 \multicolumn{2}{c}{\textbf{Humanoid \emph{run}}} \Tstrut \\
 \toprule
 \midrule
 policy network & \\
 \hspace{2em} layer sizes & $(400, 400, 300)$ \\
 \hspace{2em} take $\tanh$ of action mean? & no \\
 \hspace{2em} minimum variance & $10^{-8}$ \\
 \midrule
 critic network(s) & \\
 \hspace{2em} layer sizes & $(500, 500, 400)$ \\
 \hspace{2em} take $\tanh$ of action? & no \\
 \midrule
 MPO & \\
 \hspace{2em} KL-constraint on policy mean, $\beta_{\mu}$ & $5 \plh 10^{-3}$ \\
 \hspace{2em} KL-constraint on policy covariance, $\beta_{\Sigma}$ & $10^{-6}$ \\
 \midrule
 training & \\
 \hspace{2em} Adam learning rate & $2 \plh 10^{-4}$ \\
 \hspace{2em} Adam $\epsilon$ & $10^{-8}$ \\
 \midrule
 \bottomrule
 \\
 \multicolumn{2}{c}{\textbf{Shadow Hand}} \Tstrut \\
 \toprule
 \midrule
 MPO & \\
 \hspace{2em} actions sampled per state & $30$ \\
 \hspace{2em} default $\epsilon$ & $0.01$ \\
 \midrule
 \bottomrule
 \\
 \multicolumn{2}{c}{\textbf{Sawyer}} \Tstrut \\
 \toprule
 \midrule
 MPO & \\
 \hspace{2em} actions sampled per state & $15$ \\
 \midrule
 training & \\
 \hspace{2em} target network update period & $100$ \\
 \midrule
 \bottomrule
\end{tabular}
\caption{Hyperparameters for humanoid, Shadow Hand, and Sawyer experiments that differ from the defaults in \tableref{t:MPO}.}
\label{t:MPO_domainspecific}
\end{table}

In our experiments comparing the Pareto front found by MO-MPO versus scalarized MPO, we ran MO-MPO with a range of $\epsilon_k$ settings and ran scalarized MPO with a range of weight settings, to obtain a set of policies for each algorithm. The settings for humanoid \textbf{\emph{run}} and the Shadow Hand tasks are listed in \tableref{t:eps_weight_settings}. The settings for humanoid mocap and Sawyer are specified in the main paper, in Sec. 6.3 and 6.4, respectively.

\begin{table}[t]
\small
\centering
\setlength\tabcolsep{3pt}
\begin{tabular}{ll} 
 \toprule \midrule
 \textbf{Condition} & \textbf{Settings} \Tstrut \\
 \midrule
 \multicolumn{2}{l}{Humanoid \textbf{\emph{run}} (one seed per setting)} \\
 \rule{0pt}{3ex} 
 \hspace{2em} scalarized MPO & $w_\text{task} = 1 - w_\text{penalty}$ \\
 & $w_\text{penalty} \in \text{linspace}(0, 0.15, 100)$ \\
 \rule{0pt}{2ex} 
 \hspace{2em} MO-MPO & $\epsilon_\text{task} = 0.1$ \\
 & $\epsilon_\text{penalty} \in \text{linspace}(10^{-6}, 0.15, 100)$ \\
 \midrule
 \multicolumn{2}{l}{Humanoid \textbf{\emph{run}}, normal vs. scaled (three seeds per setting)} \\
 \rule{0pt}{3ex} 
 \hspace{2em} scalarized MPO & $w_\text{task} = 1 - w_\text{penalty}$ \\
 & $w_\text{penalty} \in \{0.01, 0.05, 0.1\}$ \\
 \rule{0pt}{2ex} 
 \hspace{2em} MO-MPO & $\epsilon_\text{task} = 0.1$ \\
 & $\epsilon_\text{penalty} \in \{0.01, 0.05, 0.1\}$ \\
 \midrule
 \multicolumn{2}{l}{Shadow Hand \textbf{\emph{touch}} and \textbf{\emph{turn}} (three seeds per setting)} \\
 \rule{0pt}{3ex} 
 \hspace{2em} scalarized MPO & $w_\text{task} = 1 - w_\text{penalty}$ \\
 & $w_\text{penalty} \in \text{linspace}(0, 0.9, 10)$ \\
 \rule{0pt}{2ex} 
 \hspace{2em} MO-MPO & $\epsilon_\text{task} = 0.01$ \\
 & $\epsilon_\text{penalty} \in \text{linspace}(0.001, 0.015, 15)$ \\
 \midrule
 \multicolumn{2}{l}{Shadow Hand \textbf{\emph{orient}} (ten seeds per setting)} \\
 \rule{0pt}{3ex} 
 \hspace{2em} scalarized MPO & $w_\text{touch} = w_\text{height} = w_\text{orientation} = 1/3$ \\
 \rule{0pt}{2ex} 
 \hspace{2em} MO-MPO & $\epsilon_\text{touch} = \epsilon_\text{height} = \epsilon_\text{orientation} = 0.01$ \\
 \midrule
 \bottomrule
\end{tabular}
\caption{Settings for $\epsilon_k$ and weights. ($\text{linspace}(x, y, z)$ denotes a set of $z$ evenly-spaced values between $x$ and $y$.)}
\label{t:eps_weight_settings}
\end{table}

In the following subsections, we first give suggestions for choosing appropriate $\epsilon_k$ to encode a particular preference (\sref{sec:setting_eps}), then we describe the discrete MO-MPO used for Deep Sea Treasure (\sref{sec:dst_details}), and finally we describe implementation details for humanoid mocap (\sref{sec:humanoid_mocap_details}).

\subsection{Suggestions for Setting $\epsilon$}
\label{sec:setting_eps}
Our proposed algorithm, MO-(V-)MPO, requires practitioners to translate a desired preference across objectives to numerical choices for $\epsilon_k$ for each objective $k$. At first glance, this may seem daunting. However, in practice, we have found that encoding preferences via $\epsilon_k$ is often more intuitive than doing so via scalarization. In this subsection, we seek to give an intuition on how to set $\epsilon_k$ for different desired preferences. Recall that each $\epsilon_k$ controls the influence of objective $k$ on the policy update, by constraining the KL-divergence between each objective-specific distribution and the current policy (Sec. 4.1). We generally choose $\epsilon_k$ in the range of $0.001$ to $0.1$.

\prg{Equal Preference} When all objectives are equally important, the general rule is to set all $\epsilon_k$ to the same value. We did this in our \textbf{\emph{orient}} and humanoid mocap tasks, where objectives are aligned or mostly aligned. In contrast, it can be tricky to choose appropriate weights in linear scalarization to encode equal preferences---setting all weights equal to $1 / K$ (where $K$ is the number of objectives) is only appropriate if the objectives' rewards are of similar scales. We explored this in Sec. 5.1 and Fig. 3.

When setting all $\epsilon_k$ to the same value, what should this value be? The larger $\epsilon_k$ is, the more influence the objectives will have on the policy update step. Since the per-objective critics are learned in parallel with the policy, setting $\epsilon_k$ too high tends to destabilize learning, because early on in training, when the critics produce unreliable Q-values, their influence on the policy will lead it in the wrong direction. On the other hand, if $\epsilon_k$ is set too low, then it slows down learning, because the per-objective action distribution is only allowed to deviate by a tiny amount from the current policy, and the updated policy is obtained via supervised learning on the combination of these action distributions. Eventually the learning will converge to more or less the same policy though, as long as $\epsilon_k$ is not set too high.

\prg{Unequal Preference} When there is a difference in preferences across objectives, the \emph{relative} scale of $\epsilon_k$ is what matters. The more the relative scale of $\epsilon_k$ is compared to $\epsilon_l$, the more influence objective $k$ has over the policy update, compared to objective $l$. And in the extreme case, when $\epsilon_l$ is near-zero for objective $l$, then objective $l$ will have no influence on the policy update and will effectively be ignored. We explored this briefly in Sec. 5.1 and Fig. 4.

One common example of unequal preferences is when we would like an agent to complete a task, while minimizing other objectives---e.g., an action norm penalty, ``pain'' penalty, or wrist force-torque penalty, in our experiments. In this case, the $\epsilon$ for the task objective should be higher than that for the other objectives, to incentivize the agent to prioritize actually doing the task. If the $\epsilon$ for the penalties is too high, then the agent will care more about minimizing the penalty (which can typically be achieved by simply taking no actions) rather than doing the task, which is not particularly useful.

The scale of $\epsilon_k$ has a similar effect as in the equal preference case. If the scale of $\epsilon_k$ is too high or too low, then the same issues arise as discussed for equal preferences. If all $\epsilon_k$ increase or decrease in scale by the same (moderate) factor, and thus their relative scales remain the same, then typically they will converge to more or less the same policy. This can be seen in Fig. 5 (right), in the Deep Sea Treasure domain: regardless of whether $\epsilon_\text{time}$ is $0.01$, $0.02$, or $0.05$, the relationship between the relative scale of $\epsilon_\text{treasure}$ and $\epsilon_\text{time}$ to the treasure that the policy converges to selecting is essentially the same.

\subsection{Deep Sea Treasure}
\label{sec:dst_details}
In order to handle discrete actions, we make several minor adjustments to scalarized MPO and MO-MPO. The policy returns a categorical distribution, rather than a Gaussian. The policy is parametrized by a neural network, which outputs a vector of logits (i.e., unnormalized log probabilities) of size $|\mathcal{A}|$. The KL constraint on the change of this policy, $\beta$, is $0.001$. The input to the critic network is the state concatenated with a four-dimensional one-hot vector denoting which action is chosen (e.g., the up action corresponds to $[1, 0, 0, 0]^\top$). Critics are trained with one-step temporal-difference error, with a discount of $0.999$. Other than these changes, the network architectures and the MPO and training hyperparameters are the same as in \tableref{t:MPO}.

\subsection{Humanoid Mocap}
\label{sec:humanoid_mocap_details}

For the humanoid mocap experiments, we used the following architecture for both MO-VMPO and VMPO: for the policy, we process a concatentation of the mocap reference observations and the proprioceptive observations by a two layer MLP with 1024 hidden units per layer. This reference encoder is followed by linear layers to produce the mean and log standard deviation of a stochastic latent variable. These latent variables are then concatenated with the proprioceptive observations and processed by another two layer MLP with 1024 hidden units per layer, to produce the action distribution. For VMPO, we use a three layer MLP with 1024 hidden units per layer as the critic. For MO-VMPO we use a shared two layer MLP with 1024 hidden units per layer followed by a one layer MLP with 1024 hidden units per layer per objective. In both cases we use k-step returns to train the critic with a discount factor of $0.95$. \tableref{t:mocap_hypers} shows additional hyperparameters used in our experiments. 
\begin{table}[t]
\small
\centering
\setlength\tabcolsep{3pt}
\begin{tabular}{lc} 
 \multicolumn{2}{c}{\textbf{Humanoid Mocap}} \Tstrut \\
 \toprule
 \midrule
 Scalarized VMPO & \\
 \hspace{2em} KL-constraint on policy mean, $\beta_\mu$ & 0.1 \\
 \hspace{2em} KL-constraint on policy covariance, $\beta_\Sigma$ & $10^{-5}$ \\
 \hspace{2em} default $\epsilon$ & $0.1$ \\
 \hspace{2em} initial temperature $\eta$ & $1$ \\
 Training & \\
 \hspace{2em} Adam learning rate & $10^{-4}$\\
 \hspace{2em} batch size & $128$\\
 \hspace{2em} unroll length (for n-step return, \sref{sec:vmpo_eval}) & $32$\\
 \midrule
 \bottomrule
 \\
\end{tabular}
\caption{Hyperparameters for the humanoid mocap experiments.}
\label{t:mocap_hypers}
\end{table}

\section{Experimental Domains}
We evaluated our approach on one discrete domain (Deep Sea Treasure), three simulated continuous control domains (humanoid, Shadow Hand, and humanoid mocap), and one real-world continuous control domain (Sawyer robot). Here we provide more detail about these domains and the objectives used in each task.
\subsection{Deep Sea Treasure}

\begin{figure}[t!]
    \centering
    \includegraphics[width=0.9\linewidth]{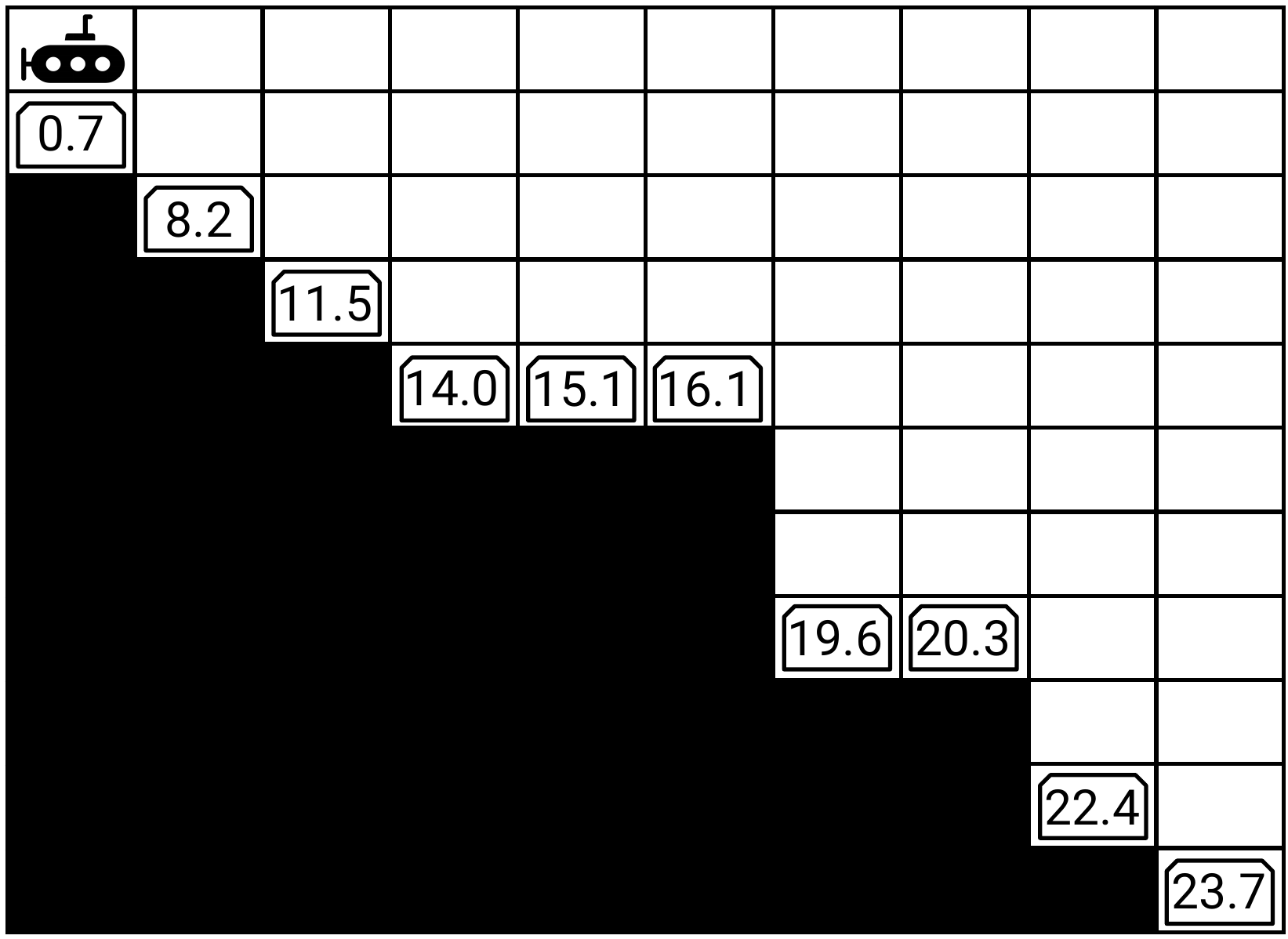}
    \caption{Deep Sea Treasure environment from \citet{Vamplew_2011}, with weights from \citet{Yang_2019}. Treasures are labeled with their respective values. The agent can move around freely in the white squares, but cannot enter the black squares (i.e., the ocean floor).}
    \label{fig:dst_env}
\end{figure}

Deep Sea Treasure (DST) is a $11 \plh 10$ grid-world domain, the state space $\mathcal{S}$ consists of the $x$ and $y$ position of the agent and the action space $\mathcal{A}$ is $\{$\texttt{up}, \texttt{right}, \texttt{down}, \texttt{left}$\}$. The layout of the environment and values of the treasures are shown in \figref{fig:dst_env}. If the action would cause the agent to collide with the sea floor or go out of bounds, it has no effect. Farther-away treasures have higher values. The episode terminates when the agent collects a treasure, or after 200 timesteps.

There are two objectives, time penalty and treasure value. A time penalty of $-1$ is given at each time step. The agent receives the value of the treasure it picks up as the reward for the treasure objective. In other words, when the agent picks up a treasure of value $v$, the reward vector is $[-1, v]$; otherwise it is $[-1, 0]$.

\subsection{Shadow Hand}
Our robot platform is a simulated Shadow Dexterous Hand \citep{ShadowHand} in the MuJoCo physics engine \citep{Todorov_2012}. The Shadow Hand has five fingers and 24 degrees of freedom, actuated by 20 motors. The observation consists of joint angles, joint velocities, and touch sensors, for a total of 63 dimensions. Each fingertip of our Shadow Hand has a $4 \plh 4$ spatial touch sensor. This sensor has three channels, measuring the normal force and the $x$ and $y$-axis tangential forces, for a sensor dimension of $4 \plh 4 \plh 3$ per fingertip. We simplify this sensor by summing across spatial dimensions, resulting in a $1 \plh 1 \plh 3$ observation per fingertip.

In the \textbf{\emph{touch}} task, there is a block in the environment that is always fixed in the same pose. In the \textbf{\emph{turn}} task, there is a dial in the environment that can be rotated, and it is initialized to a random position between $-30^{\circ}$ and $30^{\circ}$. The target angle of the dial is $0^{\circ}$. The angle of the dial is included in the agent's observation. In the \textbf{\emph{orient}} task, the robot interacts with a rectangular peg in the environment; the initial and target pose of the peg remains the same across episodes. The pose of the block is included in the agent's observation, encoded as the $xyz$ positions of four corners of the block (based on how \citet{Levine_2016} encodes end-effector pose).

\subsubsection{Balancing Task Completion and Pain}
In the \textbf{\emph{touch}} and \textbf{\emph{turn}} tasks, there are two objectives, ``pain'' penalty and task completion. A sparse task completion reward of $1$ is given for pressing a block with greater than $5$N of force or turning a dial to a fixed target angle, respectively. In both tasks, the episode terminates when the agent completes the task; i.e., the agent gets a total reward of either $0$ or $1$ for the task completion objective per episode. The Pareto plots for these two tasks in the main paper (Fig. 6) show the \emph{discounted} task reward (with a discount factor of $\gamma = 0.99$), to capture how long it takes agents to complete the task.

The ``pain'' penalty is computed as in \citep{Huang_2019}. It is based on the impact force $m(s, s')$, which is the increase in force from state $s$ to the next state $s'$. In our tasks, this is measured by a touch sensor on the block or dial. The pain penalty is equal to the negative of the impact force, scaled by how unacceptable it is:
\begin{equation}
    r_\text{pain}(s, a, s') = -\big[1 - a_\lambda(m(s,s'))\big] m(s,s') \, ,
\end{equation}
where $a_\lambda(\cdot) \in [0, 1]$ computes the acceptability of an impact force. $a_\lambda(\cdot)$ should be a monotonically decreasing function, that captures how resilient the robot and the environment are to impacts. As in \citep{Huang_2019}, we use
\begin{equation}
    a_\lambda(m) = \text{sigmoid}(\lambda_1 (-m + \lambda_2)) \, ,
\end{equation}
with $\lambda = [2, 2]^\top$. The relationship between pain penalty and impact force is plotted in \figref{fig:pain_penalty}.

\begin{figure}[t!]
    \centering
    \begin{subfigure}{0.16\textwidth}
      \centering
      \includegraphics[width=0.96\linewidth]{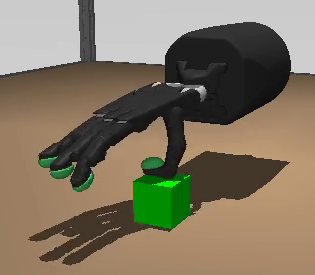}  
    \end{subfigure}%
    \begin{subfigure}{0.16\textwidth}
      \centering
      \includegraphics[width=0.96\linewidth]{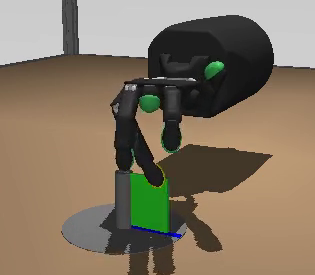}
    \end{subfigure}%
    \begin{subfigure}{0.16\textwidth}
      \centering
      \includegraphics[width=0.96\linewidth]{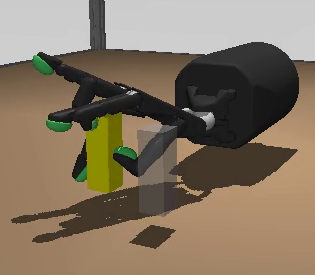}
    \end{subfigure}%
    \caption{These images show what task completion looks like for the \textbf{\emph{touch}}, \textbf{\emph{turn}}, and \textbf{\emph{orient}} Shadow Hand tasks (from left to right).}
    \label{fig:shadow}
    \vspace{-1.5em}
\end{figure}

\begin{figure}[t!]
    \centering
    \includegraphics[width=0.8\linewidth]{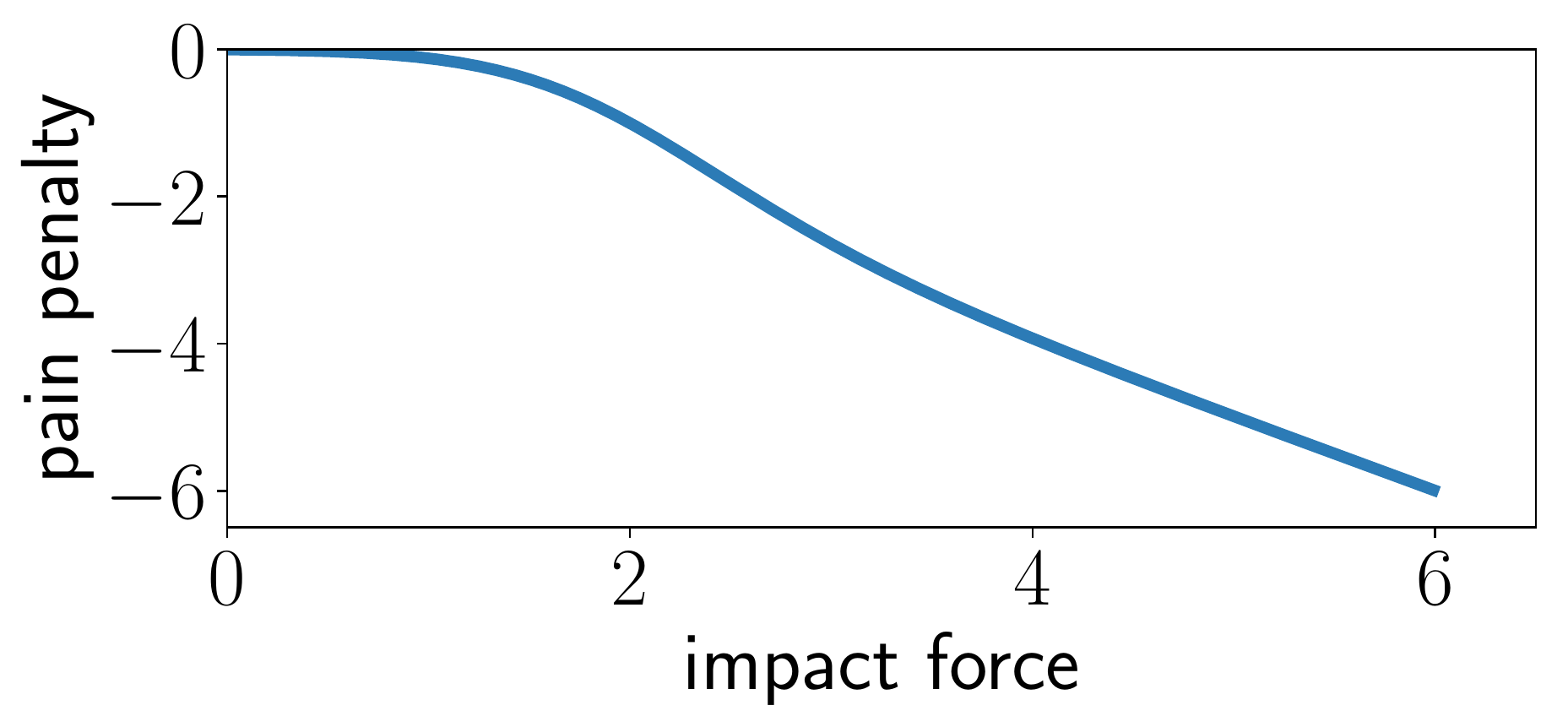}
    \caption{In the \textbf{\emph{touch}} and \textbf{\emph{turn}} Shadow Hand tasks, the pain penalty is computed based on the impact force, as plotted here. For low impact forces, the pain penalty is near-zero. For high impact forces, the pain penalty is equal to the negative of the impact force.}
    \label{fig:pain_penalty}
\end{figure}

\subsubsection{Aligned Objectives}
\begin{figure}[t!]
    \centering
    \includegraphics[width=0.8\linewidth]{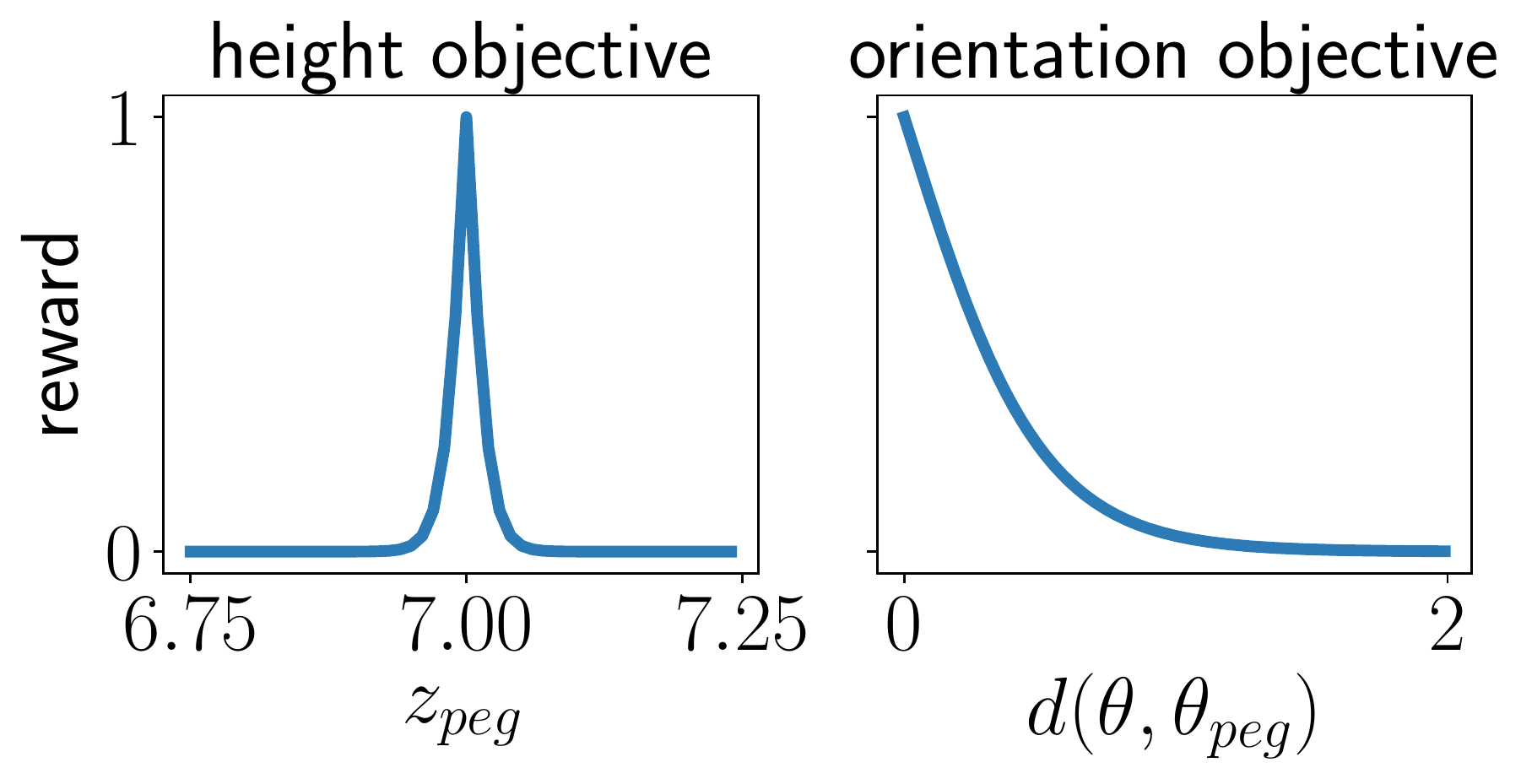}
    \caption{In the \textbf{\emph{orient}} Shadow Hand task, the height and orientation objectives have shaped rewards, as plotted here.}
    \label{fig:orient_rewards}
\end{figure}

In the \textbf{\emph{orient}} task, there are three non-competing objectives: touching the peg, lifting the peg to a target height, and orienting the peg to be perpendicular to the ground. The target $z$ position and orientation is shown by the gray transparent peg (\figref{fig:shadow}, right-most); note that the $x$ and $y$ position is not specified, so the robot in the figure gets the maximum reward with respect to all three objectives.

The touch objective has a sparse reward of $1$ for activating the peg's touch sensor, and zero otherwise. For the height objective, the target $z$ position is $7$cm above the ground; the peg's $z$ position is computed with respect to its center of mass. The shaped reward for this objective is $1 - \tanh(50 | z_\text{target} - z_\text{peg} |)$. For the orientation objective, since the peg is symmetrical, there are eight possible orientations that are equally valid. The acceptable target orientations $Q_\text{target}$ and the peg's orientation $q_\text{peg}$ are denoted as quaternions. The shaped reward is computed with respect to the closest target quaternion, as
\begin{equation*}
    \min_{q \in Q_\text{target}} 1 - \tanh(2^*d(q, q_\text{peg})) \, ,
\end{equation*}
where $d(\cdot)$ denotes the $\ell2$-norm of the axis-angle equivalent (in radians) for the distance between the two quaternions. \figref{fig:orient_rewards} shows the shaping of the height and orientation rewards.

\subsection{Humanoid}
We make use of the humanoid \textbf{\emph{run}} task from \citet{Tassa_2018}.\footnote{This task is available at \href{github.com/deepmind/dm_control}{github.com/deepmind/dm\_control}.} The observation dimension is $67$ and the action dimension is $21$. Actions are joint accelerations with minimum and maximum limits of $-1$ and $1$, respectively.
For this task there are two objectives: 
\begin{itemize}
\item The original task reward given by the environment. The goal is is to achieve a horizontal speed of $10$ meters per second, in any direction. This reward is shaped: it is equal to $\min(h/10, 1)$ where $h$ is in meters per second. For this objective we learn a Q-function.
\item Limiting energy usage, by penalizing taking high-magnitude actions. The penalty is the $\ell2$-norm of the action vector, i.e, $r_{\textrm{penalty}}(\Vec{a}) = -\|\Vec{a}\|_2$. For this objective, we do not learn a Q-function. Instead, we compute this penalty directly to evaluate a given action in a {\it state-independent way} during policy optimization.
\end{itemize}

\subsection{Humanoid Mocap}
The motion capture tracking task used in this paper was developed as part of a concurrent submission \cite{anonymous2020tracking} and does not constitute a contribution of this paper. We use a simulated humanoid adapted from the ``CMU humanoid'' available at {\color{blue} \href{https://github.com/deepmind/dm_control/tree/master/dm_control/locomotion}{dm\_control/locomotion}} \cite{merel2018hierarchical}. We adjusted various body and actuator parameters to be more comparable to that of an average human. This humanoid has 56 degrees of freedom and the observation is 1021-dimensional.

This motion capture tracking task is broadly similar to previous work on motion capture tracking \cite{chentanez2018physics,Peng_2018, merel2018hierarchical}. The task is associated with an underlying set of motion capture clips. For this paper we used roughly 40 minutes of locomotion motion capture clips from the CMU motion capture database.\footnote{This database is available at \href{http://mocap.cs.cmu.edu}{mocap.cs.cmu.edu}.}

At the beginning of a new episode, we select a frame uniformly at random from all frames in the underlying mocap data (excluding the last 10 frames of each clip). The simulated humanoid is then initialized to the pose of the selected frame.

The observations in this environment are various proprioceptive observations about the current state of the body as well as the relative target positions and orientations of 13 different body parts in the humanoid's local frame. We provide the agent with the target poses for the next 5 timesteps.

The choice of reward function is crucial for motion capture tracking tasks. Here we consider five different reward components, each capturing a different aspect of the similarity of the pose between the simulated humanoid and the mocap target. Four of our reward components were initially proposed by \citet{Peng_2018}. The first reward component is based on the difference in the center of mass:
\begin{align}
    r_{\text{com}} = \exp\left(-10\|p_\text{com}-p_{\text{com}}^{\text{ref}}\|^2\right) \, ,\nonumber
\end{align}
where $p_\text{com}$ and $p_{\text{com}}^{\text{ref}}$ are the positions of the center of mass of the simulated humanoid and the mocap reference, respectively. The second reward component is based on the difference in joint angle velocities:
\begin{align}
    r_{\text{vel}} = \exp\left(-0.1\|q_\text{vel}-q_{\text{vel}}^{\text{ref}}\|^2\right) \, ,\nonumber
\end{align}
where $q_\text{vel}$ and $q_{\text{vel}}^{\text{ref}}$ are the joint angle velocities of the simulated humanoid and the mocap reference, respectively. The third reward component is based on the difference in end-effector positions: 
\begin{align}
    r_{\text{app}} = \exp\left(-40\|p_\text{app}-p_{\text{app}}^{\text{ref}}\|^2\right) \, ,\nonumber
\end{align}
where $p_\text{app}$ and $p_\text{app}^{\text{ref}}$ are the end effector positions of the simulated humanoid and the mocap reference, respectively. The fourth reward component is based on the difference in joint orientations:
\begin{align}
    r_{\text{quat}} = \exp\left(-2\|q_{\text{quat}}\varominus q_{\text{quat}}^{\text{ref}}\|^2\right),\nonumber
\end{align}
where $\varominus$ denotes quaternion differences and $q_\text{quat}$ and $q_{\text{quat}}^{\text{ref}}$ are the joint quaternions of the simulated humanoid and the mocap reference, respectively.

Finally, the last reward component is based on difference in the joint angles and the Euclidean positions of a set of 13 body parts:
\begin{align}
    r_{\text{trunc}} = 1 - \frac{1}{0.3}(\overbrace{\|b_\text{pos}-b_\text{pos}^{\text{ref}}\|_1 + \|q_\text{pos}-q_\text{pos}^{\text{ref}}\|_1}^{:=\varepsilon}), \nonumber
\nonumber\end{align}
where $b_\text{pos}$ and $b_\text{pos}^{\text{ref}}$ correspond to the body positions of the simulated character and the mocap reference and $q_\text{pos}$, and $q_\text{pos}^{\text{ref}}$ correspond to the joint angles. 

We include an early termination condition in our task that is linked to this last reward term. We terminate the episode if $\varepsilon>0.3$, which ensures that $r_\text{trunc}\in[0,1]$.

In our MO-VMPO experiments, we treat each reward component as a separate objective. In our VMPO experiments, we use a reward of the form:
\begin{align}
    r = \frac{1}{2}r_{\text{trunc}} +
    \frac{1}{2}\left(0.1r_{\text{com}}+\lambda r_{\textbf{}{vel}}+0.15r_{\text{app}}+0.65r_{\text{quat}}\right) \, ,
\end{align}
varying $\lambda$ as described in the main paper (Sec. 6.3).

\subsection{Sawyer}
The Sawyer peg-in-hole setup consists of a Rethink Robotics Sawyer robot arm. The robot has 7 joints driven by series elastic actuators. On the wrist, a Robotiq FT 300 force-torque sensor is mounted, followed by a Robotiq 2F85 parallel gripper. We implement a 3D end-effector Cartesian velocity controller that maintains a fixed orientation of the gripper. The agent controls the Cartesian velocity setpoints. The gripper is not actuated in this setup. The observations provided are the joint positions, velocities, and torques, the end-effector pose, the wrist force-torque measurements, and the joint velocity command output of the Cartesian controller. We augment each observation with the two previous observations as a history.

In each episode, the peg is initalized randomly within an $8\plh8\plh8$ cm workspace (with the peg outside of the hole). The environment is then run at 20 Hz for 600 steps. The action limits are $\pm0.05$ m/s on each axis. If the magnitude of the wrist force-torque measurements exceeds 15 N on any horizontal axis ($x$ and $y$ axis) or 5 N on the vertical axis ($z$ axis), the episode is terminated early.

There are two objectives: the task itself (inserting the peg), and minimizing wrist force measurements.

The reward for the task objective is defined as 
\begin{align}
    r_{\text{insertion}} &= \max\Big(0.2r_{\text{approach}}, r_{\text{inserted}}\Big) \\
    r_{\text{approach}} &= s(p^{\text{peg}}, p^{\text{opening}}) \\
    r_{\text{inserted}} &= r_{\text{aligned}} \; s(p^{\text{peg}}_z, p^{\text{bottom}}_z) \\
    r_{\text{aligned}} &=  p^{\text{peg}}_{xy} - p^{\text{opening}}_{xy} < d/2
\end{align}
where $p^{\text{peg}}$ is the peg position, $p^{\text{opening}}$ the position of the hole opening, $p^{\text{bottom}}$ the bottom position of the hole and $d$ the diameter of the hole. 

$s(p_1, p_2)$ is a shaping operator
\begin{align}
    s(p_1, p_2) = 1 - \tanh^2( \frac{\atanh(\sqrt{l})}{\epsilon}  \|p_1-p_2 \|_2) \, ,
\end{align}
which gives a reward of $1-l$ if $p_1$ and $p_2$ are at a Euclidean distance of $\epsilon$. Here we chose $l_{\text{approach}}=0.95$, $\epsilon_{\text{approach}}=0.01$, $l_{\text{inserted}}=0.5$, and $\epsilon_{\text{inserted}}=0.04$. Intentionally, $0.04$ corresponds to the length of the peg such that $r_{\text{inserted}}=0.5$ if $p^{\text{peg}} = p^{\text{opening}}$. As a result, if the peg is in the approach position, $r_{\text{inserted}}$ is dominant over $r_{\text{approach}}$ in $r_{\text{insertion}}$ .

Intuitively, there is a shaped reward for aligning the peg with a magnitude of 0.2. After accurate alignment within a horizontal tolerance, the overall reward is dominated by a vertical shaped reward component for insertion. The horizontal tolerance threshold ensures that the agent is not encouraged to try to insert the peg from the side.

The reward for the secondary objective, minimizing wrist force measurements, is defined as
\begin{align}
    r_{\text{force}} = - \|F\|_1 \, ,
\end{align}
where $F = (F_x, F_y, F_z)$ are the forces measured by the wrist sensor.
\section{Algorithmic Details}

\subsection{General Algorithm Description}

We maintain one online network and one target network for each Q-function associated to objective $k$, with parameters denoted by $\phi_k$ and $\phi'_k$, respectively. We also maintain one online network and one target network for the policy, with parameters denoted by $\theta$ and $\theta'$, respectively. Target networks are updated every fixed number of steps by copying parameters from the online network. Online networks are updated using gradient descent in each learning iteration. Note that in the main paper, we refer to the target policy network as the old policy, $\pi_{\textrm{old}}$.

We use an asynchronous actor-learner setup, in which actors regularly fetch policy parameters from the learner and act in the environment, writing these transitions to the replay buffer. We refer to this policy as the behavior policy throughout the paper. The learner uses the transitions in the replay buffer to update the (online) Q-functions and the policy. Please see Algorithms \ref{Alg:MO-MPO-actor} and \ref{Alg:MO-MPO-learner} for more details on the actor and learner processes, respectively.

\begin{algorithm}
\caption{MO-MPO: Asynchronous Actor}\label{Alg:MO-MPO-actor}
\begin{algorithmic}[1]
\STATE \textbf{given} (N) reward functions $\{r_k(s,a)\}_{k=1}^N$, $T$ steps per episode
\REPEAT
\STATE Fetch current policy parameters $\a$ from learner
\STATE \textbf{// Collect trajectory from environment}
\STATE $\tau = \lbrace \rbrace$
\FOR{$t = 0, \dots, T$}

\STATE $a_t \sim \pi_\a(\cdot | s_t)$
\STATE \textbf{// Execute action and determine rewards}
\STATE $\vec{r} = \lbrack r_{1}(s_t, a_t), \dots, r_{N}(s_t, a_t) \rbrack$
\STATE $\tau \leftarrow \tau \cup \lbrace (s_t, a_t, \vec{r}, \pi_\a(a_t | s_t)) \rbrace$
\ENDFOR
\STATE Send trajectory $\tau$ to replay buffer
\UNTIL{end of training}
\end{algorithmic}
\end{algorithm}

\begin{algorithm}[t]
\small
\caption{MO-MPO: Asynchronous Learner}\label{Alg:MO-MPO-learner}
\begin{algorithmic}[1]
\STATE {\bf given} batch size (L), number of actions (M), (N) Q-functions, (N) preferences $\{\epsilon_k\}_{k=1}^N$, policy networks, and replay buffer $\mathcal{D}$
\STATE {\bf initialize} Lagrangians $\{\eta_k\}_{k=1}^N$ and $\nu$, target networks, and online networks such that $\pi_{{\a}'}= \pi_{\a}$ and $Q_{\phi'_k}=Q_{\phi_k}$ 
\REPEAT
\REPEAT
\STATE {\textbf{// Collect dataset} $\{s^i, a^{ij}, Q_k^{ij}\}_{i,j,k}^{L,M,N}$\textbf{, where}}
\STATE {\textbf{//} $s^{i}\sim\mathcal{D}, a^{ij}\sim\pi_{{\a}'}(a|s^i)$, \textbf{and} $Q_k^{ij} = Q_{{\phi_k}'}(s^i,a^{ij})$}
\STATE
\STATE {\bf // Compute action distribution for each objective}
\FOR{k = 1, \dots, $N$}
\STATE \scalebox{0.95}[1]{$
\delta_{\eta_k} \leftarrow \nabla_{\eta_k} \eta_k\epsilon_k+\eta_k\sum_{i}^L\frac{1}{L}\log\left(\sum_j^M\frac{1}{M}\exp\Big(\frac{Q_k^{ij}}{\eta_k}\Big)\right)
$}
\STATE {Update $\eta_k$ based on $\delta_{\eta_k}$}
\STATE $q_k^{ij} \propto \exp(\frac{Q_k^{ij}}{\eta_k})$
\ENDFOR
\STATE
\STATE {\bf // Update parametric policy with trust region}
\STATE {// sg denotes a stop-gradient}
\STATE $\delta_\pi \leftarrow -\nabla_\a \sum_i^L \sum_j^M \sum_k^N q_k^{ij}  \log \pi_{\a}(a^{ij}|s^i)$
\STATE  $\hfill + \textrm{sg}(\nu) \Big(\beta - \sum_i^L  \, \KL(\pi_{\a'}(a|s^i) \,\|\, \pi_\a(a|s^i) ) \diff s \Big)$
\STATE $\delta_\nu \leftarrow \nabla_\nu \nu \Big(\beta - \sum_i^L  \, \KL(\pi_{\a'}(a|s^i) \,\|\, \pi_{\textrm{sg}(\a)}(a|s^i) ) \diff s \Big)$
\STATE {Update $\pi_\a$ based on $\delta_\pi$}
\STATE {Update $\nu$ based on $\delta_\nu$}
\STATE
\STATE {\bf // Update Q-functions}
\FOR{k = 1, \dots, $N$}
\STATE $\delta_{{\phi}_k} \leftarrow \nabla_{\phi_k}  \sum_{(s_t, a_t) \in {\tau \sim \mathcal{D}}} \big( \hat{Q}_{\phi_k}(s_t, a_t) -
  Q_k^{\text{ret}} \big)^2$ 
\STATE \hfill  with $Q_k^{\text{ret}}$ as in Eq. \ref{eq:objective_q_value2}
\STATE {Update $\phi_k$ based on $\delta_{{\phi}_k}$}
\ENDFOR
\STATE
\UNTIL{fixed number of steps}
\STATE {\bf // Update target networks}
\STATE $\pi_{{\a}'}= \pi_{\a} ,Q_{{\phi_k}'}=Q_{\phi_k}$
\UNTIL{convergence}
\end{algorithmic}
\end{algorithm}

\subsection{Retrace for Multi-Objective Policy Evaluation}
Recall that the goal of the policy evaluation stage (Sec. 4.1) is to learn the Q-function $Q_{\phi_k}(s, a)$ for each objective $k$, parametrized by $\phi_k$. These Q-functions are with respect to the policy ${\pi_\textrm{old}}$. We emphasize that it is valid to use any off-the-shelf Q-learning algorithm, such as TD(0) \citep{Sutton_1998}, to learn these Q-functions. We choose to use Retrace \citep{Munos_2016}, described here.

Given a replay buffer $\mathcal{D}$ containing trajectory snippets $\tau = \{ (s_0, a_0, \vec{r}_0,s_1), \dots, (s_T, a_T, \vec{r}_T,s_{T+1})\}$, where $\vec{r}_t$ denotes a reward vector $\{r_k(s_t, a_t)\}_{k=1}^N$ that consists of a scalar reward for each of $N$ objectives, the Retrace objective is as follows:
\begin{equation}
    \min_{\phi_k} \, \mathbb{E}_{\tau \sim \mathcal{D}} \Big[ \big(Q^{ret}_{k}(s_{t}, a_{t}) - \hat{Q}_{\phi_k}(s_t, a_t))^2 \Big] \, ,
\label{eq:objective_q_value2}
\end{equation}
with
\begin{align*}
  & Q^{\text{ret}}_{k}(s_{t}, a_t) = \hat{Q}_{\phi'_k}(s_t, a_t) + \sum_{j=t}^T  \gamma^{j-t} \Big(\prod_{z=t+1}^j c_z \Big) \delta^j \, , \\ 
  & \delta^j =   r_k(s_j, a_j) +  \gamma V(s_{j+1})- \hat{Q}_{\phi'_k}(s_j, a_j) \, , \\
  & V(s_{j+1}) = \mathbb{E}_{\pi_{\textrm{old}}(a|s_{j+1})}
  [ \hat{Q}_{\phi'_k}(s_{j+1}, a) ] \, .
\end{align*}
The importance weights $c_k$ are defined as $$c_z = \min\left(1, \frac{\pi_{\textrm{old}}(a_z | s_z)}{b(a_z | s_z)}\right) \, ,$$ where $b(a_z | s_z)$ denotes the behavior policy used to collect trajectories in the environment. When $j=t$, we set $\Big(\prod_{z=t+1}^j c_z \Big) = 1$.

We also make use of a target network for each Q-function \citep{mnih2015human}, parametrized by ${\phi'_k}$, which we copy from the online network $\phi_k$ after a fixed number of gradient steps on the Retrace objective \eqref{eq:objective_q_value2}.

\subsection{Policy Fitting}
\label{sec:policy_fitting}
Recall that fitting the policy in the policy improvement stage (Sec. 4.2.2) requires solving the following constrained optimization:
\begin{align}
\pi_\textrm{new} = & \argmax_{\a}  \sum_{k=1}^N \int_s \mu(s) \int_a q_k(a|s)\log \pi_\a(a|s)\diff a\diff s  \nonumber \\ 
&\textrm{s.t.} \int_s \mu(s) \, \KL(\pi_\textrm{old}(a|s) \,\|\, \pi_\a(a|s) ) \diff s < \beta \, .
\end{align}
We first write the generalized Lagrangian equation, i.e.
\begin{align}
L(\a, \nu) = & \sum_{k=1}^N \int_s \mu(s) \int_a q_k(a|s)\log \pi_\a(a|s)\diff a\diff s \\
& + \nu \Big(\beta - \int_s \mu(s) \, \KL(\pi_\textrm{old}(a|s) \,\|\, \pi_\a(a|s) ) \diff s \Big), \nonumber
\end{align}
where $\nu$ is the Lagrangian multiplier.
Now we solve the following primal problem,
$$\max_{\a}\min_{\nu>0} L(\a,\nu) \, .$$
To obtain the parameters $\a$ for the updated policy, we solve for $\nu$ and $\a$ by alternating between 1) fixing $\nu$ to its current value and optimizing for $\a$, and 2) fixing $\a$ to its current value and optimizing for $\nu$. This can be applied for any policy output distribution.

For Gaussian policies, we decouple the update rules to optimize for mean and covariance independently, as in \cite{Abdolmaleki_2018}. This allows for setting different KL bounds for the mean ($\beta_{\mu}$) and covariance ($\beta_{\Sigma}$), which results in more stable learning. To do this, we first separate out the following two policies for mean and covariance,
\begin{align}
\pi_{\a}^{\mu}(a|s) = \mathcal{N} \Big(a; \mu_{\a}(s),\Sigma_{\a_{old}}(s) \Big) \, , \\
\pi_{\a}^{\Sigma}(a|s) = \mathcal{N} \Big(a; \mu_{\a_{old}}(s),\Sigma_{\a}(s) \Big) \, .
\end{align}
Policy $\pi_{\a}^{\mu}(a|s)$ takes the mean from the online policy network and covariance from the target policy network, and policy $\pi_{\a}^{\Sigma}(a|s)$ takes the mean from the target policy network and covariance from online policy network. Now our optimization problem has the following form:

\begin{align}
&\max_{\a}  \sum_{k=1}^N \int_s \mu(s) \int_a q_k(a|s)\Big(\log \pi_{\a}^{\mu}(a|s) \pi_{\a}^{\Sigma}(a|s) \Big)\diff a\diff s  \nonumber \\ 
&\textrm{s.t.} \int_s \mu(s) \, \KL(\pi_\textrm{old}(a|s) \,\|\, \pi_{\a}^{\mu}(a|s) ) \diff s < \beta_{\mu} \nonumber \\
 & \;\;\;\; \int_s \mu(s) \, \KL(\pi_\textrm{old}(a|s) \,\|\, \pi_{\a}^{\Sigma}(a|s) ) \diff s < \beta_{\Sigma} \, .
\end{align}

As in \cite{Abdolmaleki_2018}, we set a much smaller bound for covariance than for mean, to keep the exploration and avoid premature convergence. We can solve this optimization problem using the same general procedure as described above. 

\subsection{Derivation of Dual Function for $\eta$}

Recall that obtaining the per-objective improved action distributions $q_k(a|s)$ in the policy improvement stage (Sec. 4.2.1) requires solving a convex dual function for the temperature $\eta_k$ for each objective. For the derivation of this dual function, please refer to Appendix D.2 of the original MPO paper \cite{Abdolmaleki_2018}.

\section{MO-V-MPO: Multi-Objective On-Policy MPO} 

In this section we describe how MO-MPO can be adapted to the on-policy case, in which a state-value function $V(s)$ is learned and used to estimate the advantages for each objective. We call this approach MO-V-MPO.

\subsection{Multi-objective policy evaluation in the on-policy setting}
\label{sec:vmpo_eval}
In the on-policy setting, to evaluate the previous policy $\pi_\text{old}$, we use advantages $A(s,a)$ estimated from a learned state-value function $V(s)$, instead of a state-action value function $Q(s,a)$ as in the main text. We train a separate $V$-function for each objective by regressing to the standard $n$-step return~\cite{Sutton_1998} associated with each objective. More concretely, given trajectory snippets $\tau = \{ (s_0, a_0, \vec{r}_0), \dots, (s_T, a_T, \vec{r}_T)\}$ where $\vec{r}_t$ denotes a reward vector $\{r_k(s_t, a_t)\}_{k=1}^N$ that consists of rewards for all $N$ objectives, we find value function parameters $\phi_k$ by optimizing the following objective:
\begin{align}
\min_{\phi_{1:k}} \sum_{k=1}^N \mathbb{E}_{\tau} \Big[ \big( G_k^{(T)}(s_t,a_t) - V^{\pi_\textrm{old}}_{\phi_k}(s_t))^2 \Big].
\end{align}
Here $G_k^{(T)}(s_t,a_t)$ is the $T$-step target for value function $k$, which uses the actual rewards in the trajectory and bootstraps from the current value function for the rest: $G^{(T)}_k(s_t,a_t)=\sum_{\ell=t}^{T-1} \gamma^{\ell-t}r_k(s_\ell,a_\ell) + \gamma^{T-t}V^{\pi_\text{old}}_{\phi_k}(s_{\ell+T})$. The advantages are then estimated as $A^{\pi_\text{old}}_k(s_t,a_t)=G^{(T)}_k(s_t,a_t) - V^{\pi_\text{old}}_{\phi_k}(s_t)$.

\subsection{Multi-objective policy improvement in the on-policy setting}

Given the previous policy $\pi_\text{old}(a|s)$ and estimated advantages $\{ A^{\pi_\text{old}}_k(s,a) \}_{k=1,\ldots,N}$ associated with this policy for each objective, our goal is to improve the previous policy. To this end, we first learn an improved variational distribution $q_k(s,a)$ for each objective, and then combine and distill the variational distributions into a new parametric policy $\pi_\text{new}(a|s)$. Unlike in MO-MPO, for MO-V-MPO we use the joint distribution $q_k(s,a)$ rather than local policies $q_k(a|s)$ because, without a learned $Q$-function, only one action per state is available for learning. This is a multi-objective variant of the two-step policy improvement procedure employed by V-MPO~\cite{Song2020V-MPO}.

\subsubsection{Obtaining improved variational distributions per objective (Step 1)}

In order to obtain the improved variational distributions $q_k(s,a)$, we optimize the RL objective for each objective $A_{k}$:
\begin{align}
\label{eq:vmpo-e-step}
&\max_{q_{k}} \int_{s,a} \, q_k(s,a) \, A_{k}(s,a) \diff a \diff s\\
&\textrm{s.t. } \textrm{KL}(q_k(s,a) \| p_\textrm{old}(s,a)) < \epsilon_k \, , \nonumber
\end{align}
where the KL-divergence is computed over all $(s,a)$, $\epsilon_k$ denotes the allowed expected KL divergence, and $p_\text{old}(s,a)=\mu(s)\pi_\text{old}(a|s)$ is the state-action distribution associated with $\pi_\text{old}$.

As in MO-MPO, we use these $\epsilon_k$ to define the preferences over objectives. More concretely, $\epsilon_k$ defines the allowed contribution of objective $k$ to the change of the policy. Therefore, the larger a particular $\epsilon_k$ is with respect to others, the more that objective $k$ is preferred. On the other hand, if $\epsilon_{k} = 0$ is zero, then objective $k$ will have no contribution to the change of the policy and will effectively be ignored.

Equation \eqref{eq:vmpo-e-step} can be solved in closed form:
\begin{equation}
    q_k(s,a) \propto p_\textrm{old}(s,a) \exp\Big(\frac{A_k(s,a)}{\eta_k}\Big) \, ,
\end{equation} where the temperature $\eta_k$ is computed based on the constraint $\epsilon_k$ by solving the following convex dual problem
\begin{align}
\eta_k = &\argmin_{{\eta_k}}\bigg[
\eta_k \, \epsilon_k \, + \\ &\eta_k \log \int_{s,a} p_\textrm{old}(s,a)\exp\Big(\frac{A_k(s,a)}{\eta}\Big) \diff a \diff s\bigg] \, . \nonumber
\end{align} 
We can perform the optimization along with the loss by taking a gradient descent step on $\eta_k$, and we initialize with the solution found in the previous policy iteration step. Since $\eta_k$ must be positive, we use a projection operator after each gradient step to maintain $\eta_k>0$.

\prg{Top-$k$ advantages} As in \citet{Song2020V-MPO}, in practice we used the samples corresponding to the top 50\% of advantages in each batch of data.

\subsubsection{Fitting a new parametric policy (Step 2)}

We next want to combine and distill the state-action distributions obtained in the previous step into a single parametric policy $\pi_\text{new}(a|s)$ that favors all of the objectives according to the preferences specified by $\epsilon_k$. For this we solve a supervised learning problem that fits a parametric policy as follows:

\begin{align}
\pi_\textrm{new} = & \argmax_{\theta}  \sum_{k=1}^N \int_{s,a} q_k(s,a)\log \pi_\theta(a|s)\diff a\diff s  \nonumber \\ 
&\textrm{s.t.} \int_s \, \KL(\pi_\textrm{old}(a|s) \,\|\, \pi_\theta(a|s) ) \diff s < \beta \, ,
\label{eq:vmpo-m-step}
\end{align}

where $\theta$ are the parameters of our function approximator (a neural network), which we initialize from the weights of the previous policy $\pi_\textrm{old}$, and the KL constraint enforces a trust region of size $\beta$ that limits the overall change in the parametric policy, to improve stability of learning. As in MPO, the KL constraint in this step has a regularization effect that prevents the policy from overfitting to the local policies and therefore avoids premature convergence.  

In order to optimize Equation~\eqref{eq:vmpo-m-step}, we employ Lagrangian relaxation similar to the one employed for ordinary V-MPO~\cite{Song2020V-MPO}.

\section{Multi-Objective Policy Improvement as Inference}
In the main paper, we motivated the multi-objective policy update rules from an intuitive perspective. In this section, we show that our multi-objective policy improvement algorithm can also be derived from the RL-as-inference perspective. In the policy improvement stage, we assume that a $Q$-function for each objective is given, and we would like to improve our policy with respect to these $Q$-functions. The derivation here extends the derivation for the policy improvement algorithm in (single-objective) MPO in \citet{abdolmaleki2018relative} (in appendix) to the multi-objective case.

We assume there are observable binary improvement events, $\{R_k\}_{k=1}^N$, for each objective. $R_k=1$ indicates that our policy has improved for objective $k$, while $R_k=0$ indicates that it has not. Now we ask, if the policy has improved with respect to \emph{all} objectives, i.e., $\{R_k=1\}_{k=1}^N$, what would the parameters $\a$ of that improved policy be? More concretely, this is equivalent to the maximum a posteriori estimate for $\{R_k = 1\}_{k=1}^N$:
\begin{align}
\max_\a p_{\a}(R_1 =1, R_2=1, \dots, R_N=1) \, p(\a) \, ,
\end{align}
where the probability of improvement events depends on $\a$. Assuming independence between improvement events $R_k$ and using log probabilities leads to the following objective:
\begin{align}
\max_\a \sum_{k=1}^N\log p_{\a}(R_k =1) + \log p(\a) \, .
\label{eq:map}
\end{align}
The prior distribution over parameters, $p(\a)$, is fixed during the policy improvement step. We set the prior such that $\pi_\a$ stays close to the target policy during each policy improvement step, to improve stability of learning (\sref{sec:m_step}).

We use the standard expectation-maximization (EM) algorithm to efficiently maximize $\sum_{k=1}^N\log p_{\a}(R_k =1)$. The EM algorithm repeatedly constructs a tight lower bound in the E-step, given the previous estimate of $\a$ (which corresponds to the target policy in our case), and then optimizes that lower bound in the M-step. We introduce a variational distribution $q_k(s, a)$ per objective, and use this to decompose the log-evidence $\sum_{k=1}^N\log p_{\a}(R_k =1)$ as follows:
\begin{align}
\label{eq:decomp}
\sum_{k=1}^N & \log p_{\a}(R_k =1) \\
& = \sum_{k=1}^N \KL(q_k(s,a)\,\|\,p_{\a}(s,a|R_k=1)) \, - \nonumber \\
& \;\;\;\; \sum_{k=1}^N \KL(q_k(s,a)\,\|\,p_{\a}(R_k=1, s, a)) \, . \nonumber
\end{align}
The second term of this decomposition is expanded as:
\begin{align}
\sum_{k=1}^N & \KL(q_k(s,a)\,\|\,p_{\a}(R_k=1, s, a)) \\
& = \sum_{k=1}^N \mathbb{E}_{q_k(s,a)} \Big[ \log \frac{p_{\a}(R_k=1,s,a)}{q_k(s,a)} \Big] \nonumber \\
& = \sum_{k=1}^N \mathbb{E}_{q_k(s,a)} \Big[ \log \frac{p(R_k=1|s,a)\pi_\a(a|s)\mu(s)}{q_k(s,a)}\Big] \, , \nonumber
\end{align}
where $\mu(s)$ is the stationary state distribution, which is assumed to be given in each policy improvement step. In practice, $\mu(s)$ is approximated by the distribution of the states in the replay buffer.
$p(R_k=1|s,a)$ is the likelihood of the improvement event for objective $k$, if our policy chose action $a$ in state $s$.

The first term of the decomposition in Equation \eqref{eq:decomp} is always positive, so the second term is a lower bound on the log-evidence $\sum_{k=1}^N\log p_{\a}(R_k =1)$. $\pi_\a(a|s)$ and $q_k(a|s) = \frac{q_k(s,a)}{\mu(s)}$ are unknown, even though $\mu(s)$ is given. 
In the E-step, we estimate $q_k(a|s)$ for each objective $k$ by minimizing the first term, given $\a = \a'$ (the parameters of the target policy). Then in the M-step, we find a new $\a$ by fitting the parametric policy to the distributions from first step.

\subsection{E-Step}
The E-step corresponds to the first step of policy improvement in the main paper (Sec. 4.2.1). In the E-step, we choose the variational distributions $\{q_k(a|s)\}_{k=1}^N$ such that the lower bound on $\sum_{k=1}^N\log p_{\a}(R_k =1)$ is as tight as possible when $\a=\a'$, the parameters of the target policy. The lower bound is tight when the first term of the decomposition in Equation \eqref{eq:decomp} is zero, so we choose $q_k$ to minimize this term. We can optimize for each variational distribution $q_k$ independently:
\begin{align}
q_k(a|s) &= \argmin_{q} \mathbb{E}_{\mu(s)} \Big[ \KL(q(a|s)\,\|\,p_{\a'}(s,a|R_k=1)) \Big] \nonumber \\
&= \argmin_{q} \mathbb{E}_{\mu(s)} \Big[ \KL(q(a|s)\,\|\,\pi_{\a'}(a|s)) \, -  \nonumber \\
& \qquad\qquad\qquad\quad\;\; \mathbb{E}_{q(a|s)} \log p(R_k=1|s,a)) \Big] \, .
\label{eq:estep}
\end{align}

We can solve this optimization problem in closed form, which gives us

$$q_k(a|s) = \frac{\pi_{\a'}(a|s) \, p(R_k=1|s,a)}{\int \pi_{\a'}(a|s) \, p(R_k=1|s,a) \diff a} \, .$$

This solution weighs the actions based on their relative improvement likelihood $p(R_k=1|s,a)$ for each objective. We define the likelihood of an improvement event $R_k$ as $$p(R_k=1|s,a) \propto \exp\Big(\frac{Q_k(s,a)}{\alpha_k}\Big) \, ,$$ where $\alpha_k$ is an objective-dependent temperature parameter that controls how greedy the solution $q_k(a|s)$ is with respect to its associated objective $Q_k(s,a)$. For example, given a batch of state-action pairs, at the extremes, an $\alpha_k$ of zero would give all probability to the state-action pair with the maximum Q-value, while an $\alpha_k$ of positive infinity would give the same probability to all state-action pairs.
In order to automatically optimize for $\alpha_k$, we plug the exponential transformation into \eqref{eq:estep}, which leads to
\begin{align} 
q_k(a|s) & = \argmax_{q} \int \mu(s)\int q(a|s) Q_k(s,a) \diff a \diff s - \nonumber \\
& \qquad \; \alpha_k \int \mu(s) \KL(q(a|s) \,\|\, \pi_{\a'}(a|s) ) \diff s \, .
\end{align}  
If we instead enforce the bound on KL divergence as a hard constraint (which we do in practice), that leads to: 
\begin{align}
\label{eq:findingq}
& q_k(a|s) = \argmax_{q} \int \mu(s)\int q(a|s) Q_k(s,a) \diff a \diff s \\ 
& \qquad \qquad \;\; \textrm{s.t.} \int \mu(s) \KL(q(a|s) \,\|\, \pi_{\a'}(a|s) ) \diff s < \epsilon_k, \nonumber 
\end{align}
where, as described in the main paper, the parameter $\epsilon_k$ defines preferences over objectives. If we now assume that $q_k(a|s)$ is a non-parametric sample-based distribution as in \cite{Abdolmaleki_2018}, we can solve this constrained optimization problem in closed form for each sampled state $s$, by setting
\begin{equation}
q(a|s) \propto \pi_{\a'}(a|s) \exp\Big(\frac{Q_k(s,a)}{\eta_k}\Big) \, ,
\label{eq:nonparametric}
\end{equation}
where $\eta_k$ is computed by solving the following convex dual function:
\begin{align}
\label{eq:etadual}
&\eta_k=\argmin_{\eta}
\eta\epsilon_k+\\
& \qquad \eta\int\mu(s)\log\int \pi_{\a'}(a|s)\exp\Big(\frac{Q_k(s,a)}{\eta}\Big)\diff a \diff s \, . \nonumber
\end{align}
Equations \eqref{eq:findingq}, \eqref{eq:nonparametric}, and \eqref{eq:etadual} are equivalent to those given for the first step of multi-objective policy improvement in the main paper (Sec. 4.2.1). Please see refer to Appendix D.2 in \cite{Abdolmaleki_2018} for details on derivation of the dual function.

\subsection{M-Step}
\label{sec:m_step}
After obtaining the variational distributions $q_k(a|s)$ for each objective $k$, we have found a tight lower bound for $\sum_{k=1}^N \log p_\a(R_k=1)$ when $\a=\a'$. We can now obtain the parameters $\a$ of the updated policy $\pi_{\a}(a|s)$ by optimizing this lower bound. After rearranging and dropping the terms in \eqref{eq:map} that do not involve $\a$, we obtain
\begin{align*}
\a^* =& \argmax_{\a} \sum_{k=1}^N\int \mu(s) \int q_k(a|s)\log \pi_\a(a|s)\diff a \diff s \\
   & \qquad \quad + \log p(\a).
\end{align*}
This corresponds to the supervised learning step of policy improvement in the main paper (Sec. 4.2.2).
 
For the prior $p(\a)$ on the parameters $\a$, we enforce that the new policy $\pi_{\a}(a|s)$ is close to the target policy $\pi_{\a'}(a|s)$. We therefore optimize for 
\begin{align*}
&\a^* = \argmax_{\a}  \sum_{k=1}^N \int \mu(s) \int q_k(a|s)\log \pi_\a(a|s) \diff a \diff s \\
& \qquad \textrm{s.t.} \int \mu(s) \KL(\pi_{\a'}(a|s) \,\|\, \pi_{\a}(a|s) ) \diff s < \beta. \nonumber 
\end{align*}
Because of the constraint on the change of the parametric policy, we do not greedily optimize the M-step objective.  See \cite{Abdolmaleki_2018} for more details.

\end{document}